\documentclass[10pt,twocolumn,letterpaper]{article}

\usepackage[pagenumbers]{cvpr} 

\usepackage{graphicx}
\usepackage{amsmath}
\usepackage{amssymb}
\usepackage{booktabs}

\usepackage[pagebackref,breaklinks,colorlinks]{hyperref}

\usepackage{multirow}
\usepackage[table,xcdraw]{xcolor}

\usepackage[utf8]{inputenc}
\usepackage[T1]{fontenc}
\usepackage{amsmath,amsfonts,mathabx}
\usepackage{epigraph}

\usepackage{etoolbox}
\makeatletter
\newlength\epitextskip
\pretocmd{\@epitext}{\em}{}{}
\apptocmd{\@epitext}{\em}{}{}
\patchcmd{\epigraph}{\@epitext{#1}\\}{\@epitext{#1}\\[\epitextskip]}{}{}
\makeatother
\def\DatasetName{\emph{Human-Art}}

\definecolor{Red}{RGB}{192, 0, 0}
\definecolor{Blue}{RGB}{12, 114, 186}
\definecolor{Yellow}{RGB}{218, 169, 20}

\definecolor{HighlightBlue}{RGB}{0, 100, 148}
\definecolor{HighlightRed}{RGB}{230, 57, 70}

\definecolor{LightRed}{HTML}{ffe0e0}
\definecolor{LightBlue}{HTML}{def5ff}
\definecolor{LightYellow}{HTML}{FFF6DB}
\definecolor{LightGreen}{HTML}{eff9f0}

\usepackage{graphicx}

\usepackage{enumitem}
\setenumerate[1]{itemsep=1pt,partopsep=1pt,parsep=\parskip,topsep=2.5pt}
\setitemize[1]{itemsep=1pt,partopsep=1pt,parsep=\parskip,topsep=2.5pt}
\setdescription{itemsep=1pt,partopsep=1pt,parsep=\parskip,topsep=2.5pt}

\usepackage[accsupp]{axessibility}  

\usepackage{array}
\usepackage{booktabs}

\usepackage{threeparttable}

\usepackage{float}

\usepackage[capitalize]{cleveref}
\crefname{section}{Sec.}{Secs.}
\Crefname{section}{Section}{Sections}
\Crefname{table}{Table}{Tables}
\crefname{table}{Tab.}{Tabs.}

\def\cvprPaperID3638 
\def\confName{CVPR}
\def\confYear{2023}

\begin{document}

\title{Human-Art: A Versatile Human-Centric Dataset\\ Bridging Natural and Artificial Scenes}

\author{Xuan Ju$^{1,2}$\thanks{Equal contribution.}~~\thanks{Work done during an internship at IDEA.}~~, Ailing Zeng$^{1}$\footnote[1]{}~~\thanks{Corresponding author.}~~, Jianan Wang$^{1}$, Qiang Xu$^{2}$, Lei Zhang$^{1}$ \\
$^{1}$International Digital Economy Academy, $^{2}$The Chinese University of Hong Kong \\
{\tt\small \{xju22, qxu\}@cse.cuhk.edu.hk, \{zengailing, wangjianan, leizhang\}@idea.edu.cn}\\
\url{https://idea-research.github.io/HumanArt/}
}
\maketitle

\begin{abstract}
Humans have long been recorded in a variety of forms since antiquity.
For example, sculptures and paintings were the primary media for depicting human beings before the invention of cameras.
However, most current human-centric computer vision tasks like human pose estimation and human image generation focus exclusively on natural images in the real world. 
Artificial humans, such as those in sculptures, paintings, and cartoons, are commonly neglected, making existing models fail in these scenarios.

As an abstraction of life, art incorporates humans in both natural and artificial scenes.
We take advantage of it and introduce the \textbf{Human-Art} dataset to bridge related tasks in natural and artificial scenarios. 
Specifically, Human-Art contains $50k$ high-quality images with over $123k$ person instances from $5$ natural and $15$ artificial scenarios, which are annotated with bounding boxes, keypoints, self-contact points, and text information for humans represented in both 2D and 3D. 
It is, therefore, comprehensive and versatile for various downstream tasks. 
We also provide a rich set of baseline results and detailed analyses for related tasks, including human detection, 2D and 3D human pose estimation, image generation, and motion transfer. As a challenging dataset, we hope Human-Art can provide insights for relevant research and open up new research questions.
\end{abstract}


\setlength{\epigraphwidth}{8cm}

\vspace{-0.5cm}

\section{Introduction}
\label{sec:introduction}

\vspace{-0.3cm}

    \epigraph{" Art is inspired by life but \emph{beyond} it."}{}

    \vspace{-0.7cm}

    \begin{figure*}[ht]
    \centering
    \includegraphics[width=0.9\linewidth]{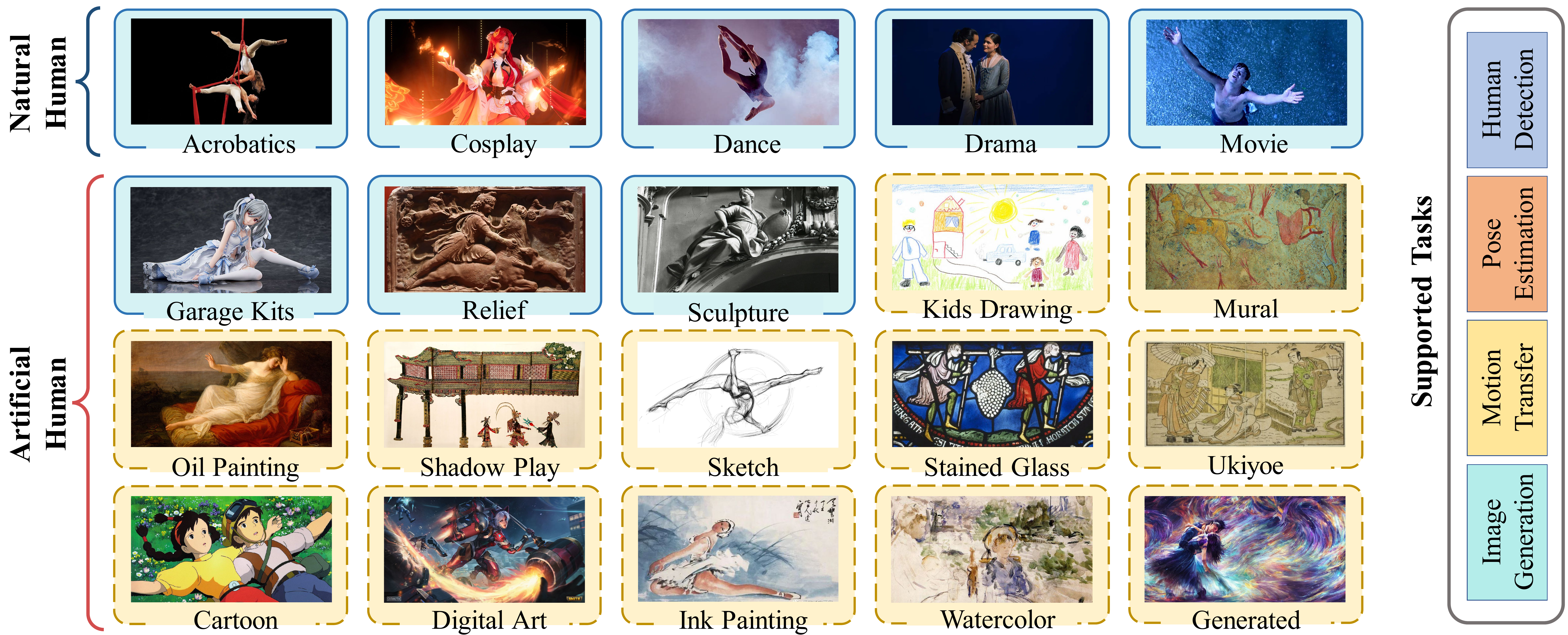}
    \vspace{-3pt}
    \caption{\DatasetName ~is a versatile human-centric dataset to bridge the gap between natural and artificial scenes. It includes $20$ high-quality scenes, including natural and artificial humans in both 2D representation (yellow dashed boxes) and 3D representation (blue solid boxes).
    } 
    \label{fig:main}
    \vspace{-10pt}
    \end{figure*}

   Human-centric computer vision (CV) tasks such as human detection~\cite{narasimhaswamy2022whose}, pose estimation~\cite{liu2022recent}, motion transfer~\cite{tao2022structure}, and human image generation~\cite{men2020controllable} have been intensively studied and achieved remarkable performances in the past decade, thanks to the advancement of deep learning techniques. Most of these works use  datasets~\cite{lin2014microsoft,h36m_pami,li2019crowdpose,narasimhaswamy2022whose} 
   that focus on humans in natural scenes captured by cameras due to the practical demands and easy accessibility.

    However, besides being captured by cameras, humans are frequently present and recorded in various other forms, 
    from ancient murals on walls to portrait paintings on paper to computer-generated virtual figures in digital form. 
    However, existing state-of-the-art (SOTA) human detection and pose estimation models~\cite{zhang2022dino,xu2022vitpose} trained on commonly used datasets such as MSCOCO~\cite{lin2014microsoft} generally work poorly on these scenarios. For instance, the average precision of such models can be as high as $63.2$\% and $79.8$\% on natural scenes but drops significantly to $12.6$\% and $28.7$\% on the \emph{sculpture} scene.
    A fundamental reason is the domain gap between natural and artificial scenes. Also, the scarcity of datasets with artificial human scenes 
    significantly restricts the development of tasks such as anime character image generation~\cite{zheng2020learning,yang2022vtoonify,chen2021improving}, character rendering~\cite{lin2022collaborative}, and character motion retargeting~\cite{yang2020transmomo,aberman2020skeleton,mourot2022survey} in computer graphics and other areas. With the growing interest in virtual reality (VR), augmented reality (AR), and metaverse, this problem is exacerbated and demands immediate attention.

    There are a few small datasets incorporating humans in artificial environments in the literature.
    Sketch2Pose~\cite{brodt2022sketch2pose} and ClassArch~\cite{madhu2020enhancing} collect images in sketch and vase painting respectively. Consequently, they are only applicable to the corresponding context. People-Art~\cite{westlake2016detecting} is a human detection dataset that consists of $1490$ paintings. It covers artificial scenes in various painting styles, but its categories are neither mutually exclusive nor collectively comprehensive. More importantly, the annotation type and image number in People-Art are limited, and hence this dataset is mainly used for testing (instead of training) object detectors.

    Art presents humans in both natural and artificial scenes in various forms, e.g., dance, paintings, and sculptures.
    In this paper, we take advantage of the classification of visual arts to introduce \DatasetName, a versatile human-centric dataset, to bridge the gap between natural and artificial scenes. \DatasetName~is hierarchically structured and includes high-quality human scenes in rich scenarios with precise manual annotations.
    Specifically, it is composed of $50k$ images with about $123k$ person instances in $20$ artistic categories, including $5$ natural and $15$ artificial scenarios in both 2D and 3D, as shown in Fig.~\ref{fig:main}. 
    To support both recognition and generation tasks, \DatasetName~provides precise manual annotations containing human bounding boxes,
    2D keypoints, self-contact points, and text descriptions. It can compensate for the lack of scenarios in prior datasets (e.g., MSCOCO~\cite{lin2014microsoft}), link virtual and real worlds, and introduce new challenges and opportunities for human-centric areas.

    \DatasetName~has the following unique characteristics:
    \vspace{-0.1cm}
    \begin{itemize}
        \item \textbf{Rich scenario}: \DatasetName~
        focuses on scenes missing in mainstream datasets (e.g.,~\cite{lin2014microsoft}), which covers most human-related scenarios. Challenging human appearances, diverse contexts, and various poses largely complement the scenario deficiency of existing datasets and will open up new challenges and opportunities.
        \item \textbf{High quality}: We guarantee inter-category variability and intra-category diversity in style, author, origin, and age. The $50$k images are manually selected from $1,000$k carefully collected images using standardized data collection, filtering, and consolidating processes.
        \item \textbf{Versatile annotations}: \DatasetName~provides carefully manual annotations of 2D human keypoints, human bounding boxes, and self-contact points to support various downstream tasks. Also, we provide accessible text descriptions to enable multi-modality learning. 
    \end{itemize}

    \vspace{-0.1cm} 
    
    With \DatasetName, we conduct comprehensive experiments and analysis on various downstream tasks including human detection, human pose estimation, human mesh recovery, image generation, and motion transfer. Although training on \DatasetName~can lead to a separate 31\% and 21\% performance boost on human detection and human pose estimation, results demonstrate that human-related CV tasks still have a long way to go before reaching maturity. 

\vspace{-1pt}
\section{Related Work}
\label{sec:related_work}

\label{sec:related_work_2d_human_centric_recognition_datasets}

\vspace{-2pt}

    \begin{table*}[hbtp]
    \centering

    \begin{threeparttable}
    \begin{tabular}{ccccccccccc}
    \specialrule{0.1em}{1pt}{1pt}
    
    \rowcolor[HTML]{EFEFEF} 
       \multicolumn{1}{c}{\cellcolor[HTML]{EFEFEF}\textbf{Dataset}} & \textbf{Image} & \textbf{Instance}                            & \textbf{\begin{tabular}[c]{@{}c@{}}Keypoint\\ Number\end{tabular}} & \textbf{Bbox} & \textbf{Pose} & \textbf{\begin{tabular}[c]{@{}c@{}}Self-\\ Contact\end{tabular}} & \textbf{\begin{tabular}[c]{@{}c@{}}Natural\\ Scenario\end{tabular}} & \textbf{\begin{tabular}[c]{@{}c@{}}Artificial\\ Scenario\end{tabular}}  \\ \specialrule{0.1em}{1pt}{1pt}
    \rowcolor[HTML]{FFFFFF} 
   VOC2012\tnote{1}\,\cite{everingham2010pascal}                                               & 8,174          & 17,132                                       & -          & $\surd$             &     -                              &    -   & $\surd$                                                                &    -                                                                                                     \\ 
    \rowcolor[HTML]{FFFFFF} 
    MSCOCO\tnote{1}\,\cite{lin2014microsoft}                                                      & 66,808         & 273,469                                      & -        & $\surd$             &   -                                &      -    & $\surd$                                                                &   -    \\ 
    \rowcolor[HTML]{FFFFFF} 
    BodyHands\cite{narasimhaswamy2022whose}                                                      & 20,490         & 63,095                                      & -        & $\surd$             &   -                                &     -     & $\surd$                                                                &    -                                                                                                        \\ 

    People-Art\cite{westlake2016detecting}                                                   & 1,490          & 3,870                                        & -             & $\surd$             &     -                              &   -    & $\surd$                                                                 & $\surd$                  \\                                                                                    
    \midrule
    \rowcolor[HTML]{FFFFFF} 
    MSCOCO\tnote{2}\,\cite{lin2014microsoft}                                                       & 58,945         & 156,165                                      & 17         & $\surd$             & $\surd$                                 &           -     & $\surd$                                                                &   -                                                                                              \\  
    \rowcolor[HTML]{FFFFFF} 
    MPII\cite{andriluka20142d}                                                         & 24,920         & 40,522                                       & 16          & $\surd$             & $\surd$                                 &      -     & $\surd$                                                                &                     -                                                                               \\  
    \rowcolor[HTML]{FFFFFF} 
    AI Challenger\cite{wu2019large}                                                 & 240,000        & 448,776                                      & 14         & $\surd$             & $\surd$                                 &     -    & $\surd$                                                                &  -                                                                                                      \\  
    \rowcolor[HTML]{FFFFFF} 
     CrowdPose\cite{li2019crowdpose}                                                    & 20,000         & $\sim$80,000                                 & 14              & $\surd$             & $\surd$                                 &   -     & $\surd$                                                                &     -                                                                                               \\  
    \rowcolor[HTML]{FFFFFF} 
    OCHuman\cite{zhang2019pose2seg}                                                      & 4,731          & 8,110                                        & 17             & $\surd$             & $\surd$                                 &    -    & $\surd$                                                                &           -                                                                                          \\  
    \rowcolor[HTML]{FFFFFF} 
    PoseTrack\tnote{3}\, \cite{andriluka2018posetrack}                                                   & 66,374         & 153,615                                      & 15           & $\surd$             & $\surd$                                 &     -       & $\surd$                                                                & -                                                                                                  \\  
    \rowcolor[HTML]{FFFFFF} 
     HiEve\tnote{3}\, \cite{lin2020human}                                                   & 49,820         & 1,099,357                                      & 14            & $\surd$             & $\surd$                                 &   -   & $\surd$                                                                &      -                                                                                                  \\  
         \midrule

     \rowcolor[HTML]{FFFFFF} 
     ClassArch\cite{madhu2020enhancing}&1513&1728&17&$\surd$&$\surd$&-&-&$\surd$\\
     Sketch2Pose\cite{brodt2022sketch2pose}                                            & 808            & 14,772                                       & 18              & $\surd$             & $\surd$                                 &    $\surd$       &                                                     -            & $\surd$         \\ 

    \midrule 
    \rowcolor[HTML]{FFFFFF}   
    \textbf{\DatasetName~(Ours)}                                              & 50,000         & 123,131 & 21                 & $\surd$             & $\surd$                                 & $\surd$     & $\surd$                                                                & $\surd$   \\ \specialrule{0.1em}{1pt}{1pt}
    \end{tabular}
    \begin{tablenotes}
            \footnotesize
            \item[1] 
            Only calculate statistics of images that contain human bounding box annotation for detection;
            \item[2] Only calculate statistics of images that contain human keypoint annotation for human pose estimation. 
            \item[3] Video-based datasets.
    \end{tablenotes}
    \end{threeparttable}
    \vspace{-0.1cm}
    \caption{Comparison of human-centric recognition datasets, including human detection and pose estimation tasks.}
    \label{tab:human_centric_dataset}
    \vspace{-0.5cm}
    \end{table*}

   \textbf{Human-centric datasets with natural scenes:} The main tasks in human-centric recognition are human detection and pose estimation. 
    As summarized in Tab.~\ref{tab:human_centric_dataset}, most existing datasets~\cite{everingham2010pascal,andriluka20142d,wu2019large,li2019crowdpose,zhang2019pose2seg,andriluka2018posetrack,lin2020human,narasimhaswamy2022whose} annotate humans in natural scenes with bounding boxes and keypoints. 
    Among them, MSCOCO~\cite{lin2014microsoft} is most widely used due to its diverse poses and complex scenes. Numerous deep models trained with it demonstrate high performance on various downstream tasks~\cite{jin2020wholebody,xu2022vitpose,li2021rle,GyeongsikMoon2020hand4whole}.
    Pedestrian detection datasets~\cite{xu2019,sindagi2020jhu-crowd++,lim2014crowd} can also be categorized as special human detection datasets focusing on small and hazy persons in congested situations.

    Although widely used in computer vision tasks, exclusively focusing on natural scenes make models trained on these datasets fail in the artificial scenario.

    \textbf{Human-centric datasets with artificial scenes:} Only a few small-scale datasets~\cite{brodt2022sketch2pose,westlake2016detecting,madhu2020enhancing} involve artificial scenarios.
    Specifically, Sketch2Pose\cite{brodt2022sketch2pose} focuses on the \emph{sketch} scenario, and ClassArch~\cite{madhu2020enhancing} only includes ancient vase paintings. 
    People-Art~\cite{westlake2016detecting} contains both natural images and artificial images. It directly borrows the artistic painting styles from \emph{wikiart}\footnote{\url{https://www.wikiart.org/}} for categorizing artificial scenes. Inter-category similarity causes confusion, especially when the images are not manually reviewed. Also, artworks beyond paintings (e.g., sculptures and digital arts) are ignored. 
    More importantly, with the limited number of images and merely human bounding-box annotation, People-Art only supports small-scale human detection tasks.

    \textbf{Human-centric synthetic datasets:} Various synthetic human body datasets~\cite{khungurn2016pose,varol17_surreal,Patel:CVPR:2021,kadish2021improving} are proposed in the literature. However, 
    they are far less developed compared to artificial human face datasets~\cite{zheng2020cartoon,kim2021animeceleb}. Generally speaking, these datasets face the problems of unnatural interactions between background and humans and the lack of character diversities. For example, the anime/manga character dataset in~\cite{khungurn2016pose} contains 
    only $2631$ different characters despite having more than a million images.

    To sum up, it is essential and urgent to bridge the gap between natural and artificial scenes for human-centric computer vision tasks. This motivates \DatasetName, a rich-scenario human-centric dataset containing sufficient high-quality images and versatile annotations,. 

\vspace{-1pt}
\section{The \DatasetName~ Dataset}
\label{sec:the_human_art_dataset}
\vspace{-2pt}

\subsection{The Hierarchical Category Classification}
\label{sec:the_human_art_dataset_definition}

    As an abstraction of the natural world, art is a metaphorical expression of how people perceive the world.  

    This makes art a good point of penetration for scenes containing both natural and artificial scenarios.

    According to~\cite{shiner2003invention}, artistic presentations can be divided into eight classes: literature, music, architecture, painting, sculpture, drama, dance, and movies.
    Among them, painting, sculpture, drama, dance, and movies can be expressed in the form of images.
    To increase \textbf{inter-category variability}, as shown in Fig.~\ref{fig:main}, we further divide these classes into twenty categories wherein humans frequently appear: 
    \begin{itemize}
        \item 5 types of \emph{natural} human scenes: Acrobatics, Cosplay, Dance, Drama, and Movie;
        \item 3 types of \emph{3D artificial} human scenes: Garage Kits, Relief, Sculpture;
        \item 12 types of \emph{2D artificial} human scenes: Kids Drawing, Mural, Oil Painting, Shadow Play, Sketch, Stained Glass, Ukiyoe, Cartoon, Digital Art, Ink Painting, Watercolor, and Generated Images;
    \end{itemize}

    Compared to previous art-related models 
    ~\cite{PeiyuanLiao2022TheAD,karayev2013recognizing,garcia2018read} that directly borrow classification criteria from some websites such as \emph{wikiart} without examination, our classification criteria is more suitable for human-centric scenes. On the one hand, it enables us to 
    easily collect a larger amount of high-quality human-related images. On the other hand,
    the inter-category variability of \DatasetName~is significant to achieve diverse images with negligible classification errors.

    \begin{figure}[t]
    \centering
    \includegraphics[width=1.\linewidth]{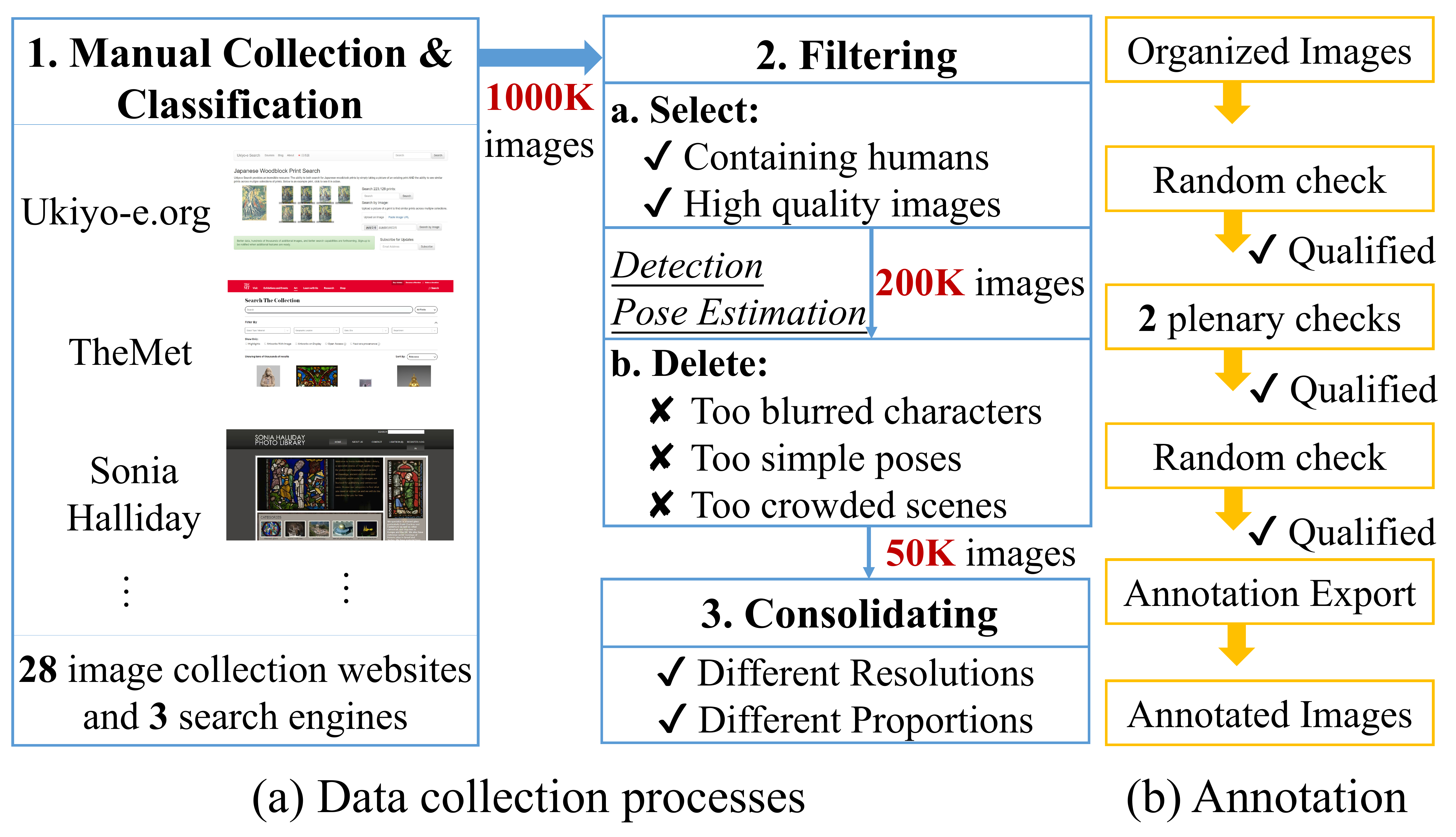}
    \caption{Data collection and annotation processes.}

    \vspace{-0.5cm}
    \label{fig:collection_process}

    \end{figure}

\subsection{Data Collection}
\label{sec:the_human_art_dataset_data_collection_and_organization}

    We design a standardized pipeline for high-quality image collection, including manual image collection \& classification, filtering, and consolidating, as shown in Fig.~\ref{fig:collection_process}.

    \begin{itemize}
        \item \textbf{Manual collection \& classification:}
        All images in \DatasetName~are either manually selected from 27 image websites that provide high-resolution images, self-collected from offline exhibitions, or generated by the popular models (e.g., Stable Diffusion ~\cite{latentdiffusion}) to ensure high quality.
        We have carefully examined a large number of possible image sources and finally determined to use 27 high-quality image websites such as ukiyo-e\footnote{\url{https://ukiyo-e.org/}}  
        and Sonia Halliday Photo Library\footnote{\url{http://www.soniahalliday.com/index.html}}.
        To ensure \textbf{intra-category diversity}, each category in \DatasetName~comes from multiple websites. 
        For search-based image websites that are not precisely aligned with our classification, we conduct searching with multiple keywords.
        We also add images from Google and Bing searches to further increase the diversity.
        Finally, we collect around one million well-classified images with the above procedure.

        \item \textbf{Filtering:}
        Despite the above efforts, the quality of images crawled from the Internet is not fully guaranteed, and a large portion of these images do not contain human beings. To tackle the above issues, we manually screen the collected images twice, with each image examined by at least two people, obtaining around $200k$ high-quality images that include humans. 
        Next, we apply a variety of human detection and pose estimation algorithms to these images. The objective is to identify and remove those images containing characters that are too simple to improve model performance, blurred characters, and crowded characters that are hard to label. This filtering step finally yields $50k$ images.
        \item \textbf{Consolidating:} We further consolidate the whole dataset with multiple resolutions: $512\times 512$, $256\times 192$, $32\times 32$, and the original resolution, for ease of use in various kinds of downstream tasks.

    \end{itemize}

    Finally, we split the \DatasetName~dataset into training, validation, and testing sets with a ratio of 70\%, 10\%, and 20\%, resulting in $35k$, $5k$, and $10k$ images in each group.

    \begin{figure}[h]
    \centering
    \includegraphics[width=1\linewidth]{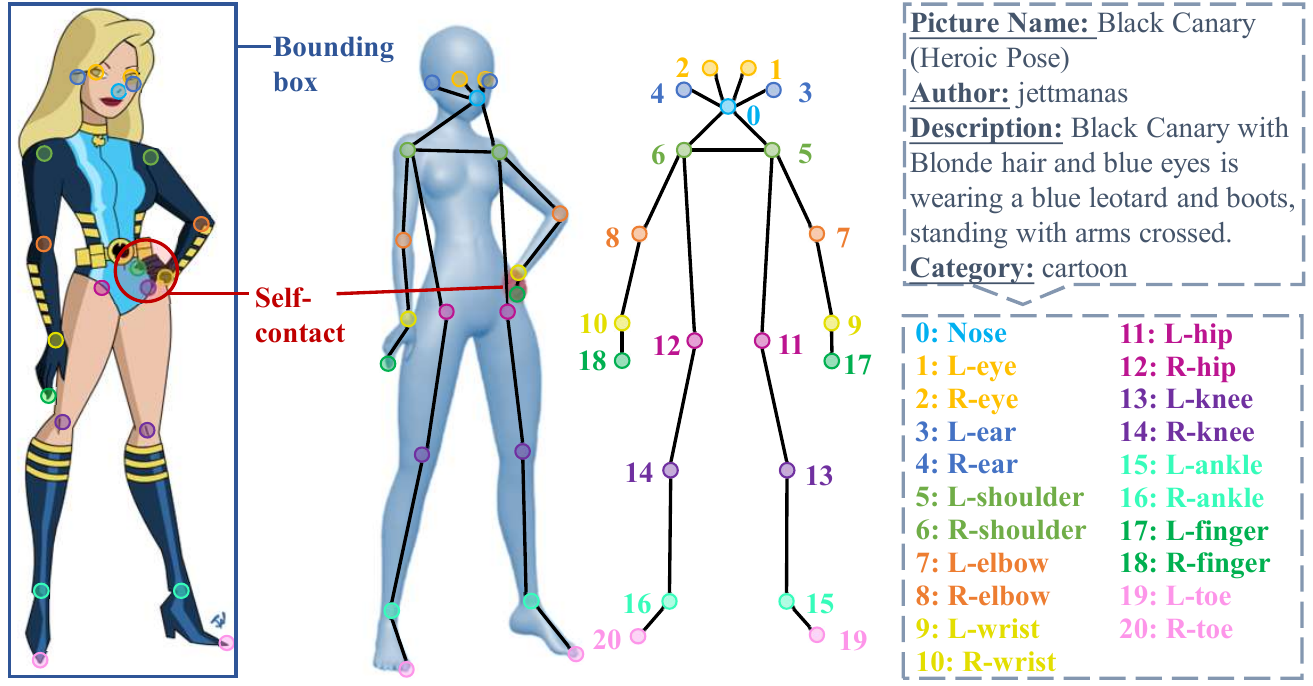}
    \caption{Illustration of the provided annotations including 2D keypoints, bounding box, self-contact point, and text description.}
    \label{fig:annotation_example}
    \vspace{-0.5cm}
    \end{figure}

\subsection{Data Annotation}
\label{sec:the_human_art_dataset_data_annotation}

    As shown in Fig.~\ref{fig:annotation_example}, \DatasetName~provides rich annotations\footnote{We do not annotate for the Generated Image category because many images in this category do not have legitimate human body parts, as elaborated in Section \ref{sec:experiments_generation}. 
    }, including human bounding-box, $21$ human keypoints (with corresponding visible/included/invisible attribute), self-contact keypoints, and text information.

    We follow MSCOCO\cite{lin2014microsoft} to define the first $17$ keypoints and add $4$ additional keypoints, left/right fingers and left/right toes, which is beneficial for 3D pose estimation and shape recovery by providing more comprehensive constraints
    ~\cite{Pavlakos_2019smplx,zeng2022smoothnet}.
    Self-contact keypoints~\cite{Muller_2021_CVPR,brodt2022sketch2pose} also demonstrate benefits for 3D pose and shape estimation by disambiguating the body part depth unknown in 2D human pose representation, avoiding self-collisions and penetrations, and being easier to use than ordinal depth\cite{pavlakos2018ordinal,sharma2019monocular,zhou2019hemlets}. 
    These annotations are especially important for humans in artworks because they suffer from more severe pose distortion and imprecise body shape due to artistic exaggeration. Self-contact keypoints are annotated as the center of the body contact surface. Although ambiguity such as distorted body proportions/shapes and incomplete/blurry human bodies exist in artworks, human annotators are capable of inferring the positions based on intuitive knowledge. Text descriptions are automatically crawled from image websites, which usually include comparatively accurate text descriptions. For images that do not contain corresponding text descriptions, we use BLIP-2~\cite{li2023blip} to generate fake labels.

    The annotations are performed by a professional team with standardized annotation and audit procedures.
    Including $35$ data annotators and $12$ data auditors, the annotation team is systematically trained before starting annotations to ensures high annotation quality and timely feedback.
    As shown in Fig.~\ref{fig:collection_process}~(b), the entire labeling process goes through two plenary quality checks and two random quality checks to ensure an accuracy of at least 98\%.

\subsection{Dataset Statistics and Analysis}
\label{sec:the_human_art_dataset_dataset_statics_and_analysis}

    \begin{figure}[h]
    \vspace{-0.4cm}
    \centering
    \includegraphics[width=0.7\linewidth]{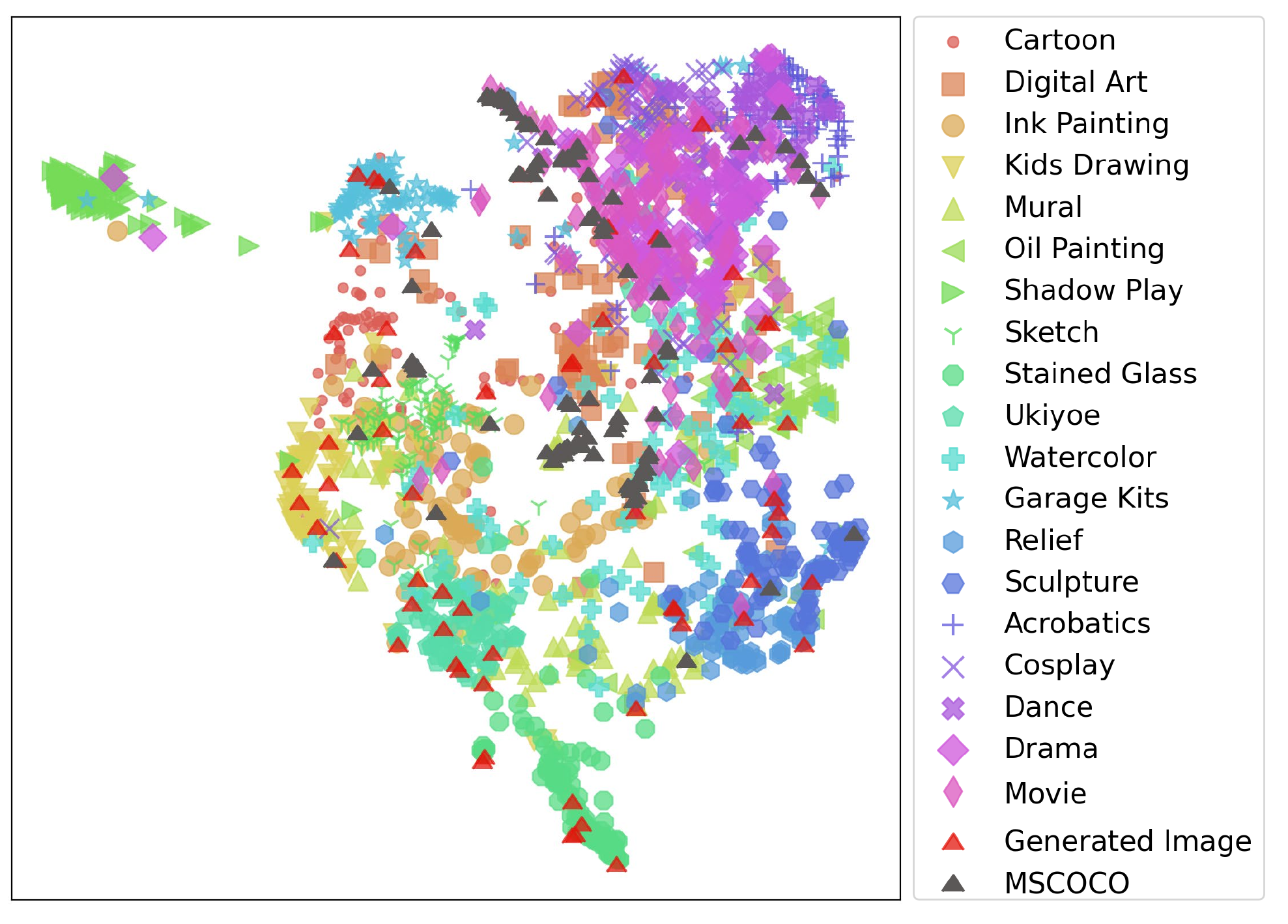}
    \caption{Illustration of the data distribution of the 20 categories in \DatasetName~and that of MSCOCO via a popular clustering method, the uniform manifold approximation and projection (UMAP)\cite{mcinnes2018umap}.}
        \vspace{-0.2cm}
    \label{fig:umap}
    \end{figure}

    \begin{table*}[htbp]
    \small
    \centering
    \setlength\tabcolsep{4pt} 
    \resizebox{0.8\textwidth}{!}{
    \begin{threeparttable}
    \begin{tabular}{ccc|cccc|cc|cc|cc}
    \toprule
    \rowcolor[HTML]{BBBBBB}  
    \multicolumn{3}{c|}{\cellcolor[HTML]{BBBBBB}\textbf{Detector}}                                                                              & \multicolumn{4}{c|}{\cellcolor[HTML]{BBBBBB}Faster R-CNN} & \multicolumn{2}{c|}{\cellcolor[HTML]{BBBBBB}YOLOX} & \multicolumn{2}{c|}{\cellcolor[HTML]{BBBBBB}Deformable DETR} & \multicolumn{2}{c}{\cellcolor[HTML]{BBBBBB}DINO} \\ \midrule
    \rowcolor[HTML]{D7D7D7} 
    \multicolumn{3}{c|}{\cellcolor[HTML]{D7D7D7}Setting}  & val  & val \tnote{*} & test  & test \tnote{*} & val    & test    & val & test & val & test\\ \midrule
    \rowcolor[HTML]{EFEFEF} 
    \multicolumn{3}{c|}{\cellcolor[HTML]{EFEFEF}MSCOCO~\cite{lin2014microsoft}}     & \textbf{52.2} & 51.6 & \textbf{-} & - &\textbf{61.9} &  \textbf{-} &  \textbf{57.2} &  \textbf{-} & \textbf{63.2} & - \\
    \cellcolor{LightGreen}  & \cellcolor{LightYellow} & Cartoon & 8.8 &37.9 & 7.0 & \underline{33.5} & 10.8  &  9.2 &  7.9 & 6.7 & 8.7 & 8.1 \\
    \cellcolor{LightGreen} & \cellcolor{LightYellow} & Digital Art & 18.8 & 46.4 & 17.8 & 44.2 & 24.1 &  22.9 &  17.6 & 15.5 & 18.6 & 18.1 \\
    \cellcolor{LightGreen} & \cellcolor{LightYellow} & Ink Painting & 11.0 & 37.7 & 9.1 & 37.2 & 15.5 &  13.0 &  11.9 & 10.0 & 14.5 & 11.6 \\
    \cellcolor{LightGreen} & \cellcolor{LightYellow} & Kids Drawing  & 6.6 & 54.2 & 8.0    & 53.6 & 6.8   &  11.5  &  5.6 & 7.2 & 6.8 & 8.2 \\
    \cellcolor{LightGreen} & \cellcolor{LightYellow} & Mural & 9.7  & 35.5 & 9.3 &  34.5  & 12.2 &  12.2  & 9.3  & 8.1   & 10.2   & 9.5  \\
    \cellcolor{LightGreen}  & \cellcolor{LightYellow} & Oil Painting & 15.9  & 41.1 & 13.7 & 37.5 & 20.8 &  18.3 &  17.1 & 14.2 & 17.0 & 15.0 \\
    \cellcolor{LightGreen}  & \cellcolor{LightYellow}  & Shadow Play & 7.5 & \textbf{64.1} & 8.2 & \textbf{63.7} & 5.4 &   7.5 &   5.3 & 5.1 & 6.4 & 7.9 \\
    \cellcolor{LightGreen} & \cellcolor{LightYellow} & Sketch &  \underline{2.6}    &  48.8 & \underline{2.4}    & 55.7  & \underline{4.6}   &  \underline{5.2}   &  5.8 & 9.2& \underline{3.6} & 7.1   \\
    \cellcolor{LightGreen} & \cellcolor{LightYellow} & Stained Glass & 8.8   & \underline{35.0} & 8.1   & 34.7 &  8.2   &  7.8  & 5.1 & \underline{4.6} & 7.8& 7.8   \\
    \cellcolor{LightGreen} & \cellcolor{LightYellow} & Ukiyoe & 12.7    & 51.9 &  12.7    & 50.3 & 13.1   & 12.8  &  8.5 & 8.4    & 11.4 & 11.4    \\
    \cellcolor{LightGreen} & \multirow{-11}{*}{\cellcolor{LightYellow}{\rotatebox{90}{2D Representation}}} & Watercolor    & 14.8   & 42.8 & 14.2   & 42.2 & 19.7   &  18.2  &  15.6 & 13.6    & 15.4 & 14.3    \\
    \cellcolor{LightGreen} & \cellcolor{LightBlue} & Garage Kits   & 22.9   & 60.0 & 22.5   & 62.5 & 22.3   &  19.9   &  17.9 & 14.6 & 22.8 & 19.5\\
    \cellcolor{LightGreen} & \cellcolor{LightBlue} & Relief & 4.9    & 37.5 & 4.7    &  33.4 & 8.4   &  9.1   &  \underline{4.7} &  5.7   & 4.4 & \underline{5.9}\\
    \multirow{-14}{*}{\cellcolor{LightGreen}\rotatebox{90}{Artificial Scene}} & \cellcolor{LightBlue} & Sculpture& 17.7   &  48.6 & 14.4   & 47.0 & 15.8  &  13.2  &   9.4 & 7.1 &  10.1 &  8.5    \\
    \cellcolor{LightRed} & \cellcolor{LightBlue} & Acrobatics    & 17.0  & 49.7 &17.0  &  53.4 & 20.0   &  19.4  &  17.3 & 17.6 & 19.4 &18.9 \\
    \cellcolor{LightRed} & \cellcolor{LightBlue} & Cosplay& 31.2   & 52.8  &\textbf{31.3}   & 56.7  & 38.0   &  \textbf{37.2}  & 34.6 & \textbf{34.5}    & 37.2 & \textbf{36.7}    \\
    \cellcolor{LightRed} & \cellcolor{LightBlue} & Dance  & 17.0   &  46.6  &18.4   & 49.3  & 20.3   & 21.1  &  17.8 & 18.5    & 19.3 & 19.6    \\
    \cellcolor{LightRed} & \cellcolor{LightBlue} & Drama  & 24.3   & 46.0 &24.8   & 48.7  & 27.4   &  27.5  &  15.4 & 25.8    & 27.8 & 16.7    \\
    \multirow{-5}{*}{\cellcolor{LightRed}\rotatebox{90}{Natural Scene}} & \multirow{-8}{*}{\cellcolor{LightBlue}\rotatebox{90}{3D Representation}}  & Movie  & 26.3   & 36.5  & 25.0   & 37.2  & 28.0   &  26.8  &  26.6 & 26.2    & 27.2 & 26.3    \\ \midrule
    \multicolumn{3}{c|}{Average}& 12.0   & 44.2  & 12.5   & 43.0  &14.4   &  14.7  &  11.7 & 11.7    & 12.6 & 12.7    \\ \bottomrule
    \end{tabular}
        \begin{tablenotes} 
            \footnotesize 
            \item[*] the baseline results we provide by training on the joint of MSCOCO~\cite{lin2014microsoft} and \DatasetName. 
    \end{tablenotes}
    \end{threeparttable}}
    \vspace{-3pt}
    \caption{Comparisons of average precision (AP) of wildly used object detection models, including Faster R-CNN~\cite{NIPS2015_14bfa6bb}, YOLOX~\cite{redmon2016you}, Deformable DETR~\cite{Zhu_detr21}, and recent SOTA DINO~\cite{zhang2022dino}. The best results are shown in \textbf{bold} and the worst results are highlighted with \underline{underlined font}. All models are trained on MSCOCO~\cite{lin2014microsoft}. Detailed settings for each model are provided in supplementary files.}
    \label{tab:human_detection}
    \vspace{-0.4cm}
    \end{table*}

   To demonstrate the inter-category variability and intra-category diversity of \DatasetName, we randomly select $100$ images from each category and also $100$ images from the MSCOCO dataset. Then we use ResNet152\cite{he2016deep} to extract image features and UMAP\cite{mcinnes2018umap} to reduce these features down to $2$ dimensions for visualization. As shown in Fig.~\ref{fig:umap}, most scenarios form their own clusters, and distinct types of images (e.g., Drama and Sculptures) lead to dramatically different distributions that are far apart. The distribution of MSCOCO is similar to the natural image categories in \DatasetName~and is included in the $20$ categories' distribution. Moreover, the generated images are scattered across the distribution as expected.

    Moreover, we calculate the invisible keypoints of \DatasetName, and the results demonstrate a higher valid keypoint percentage than the MSCOCO dataset. More dataset statistics and analysis are in the supplementary file.

\vspace{-0.1cm}
\section{Experiments}
\label{sec:experiments}

    \begin{table*}[htbp]
    \centering
        \small
    \setlength\tabcolsep{3.5pt} 
    \resizebox{0.99\textwidth}{!}{
    \begin{threeparttable}
    \begin{tabular}{ccc|cccccc|cccc|cc|cccc}
    \toprule
    \rowcolor[HTML]{BBBBBB} 
    \multicolumn{3}{c|}{\cellcolor[HTML]{BBBBBB}\textbf{Detector}}      & \multicolumn{6}{c|}{\cellcolor[HTML]{BBBBBB}\textbf{Faster R-CNN + HRNet}}  & \multicolumn{4}{c|}{\cellcolor[HTML]{BBBBBB}\textbf{YOLOX + ViTPose}} & \multicolumn{2}{c|}{\cellcolor[HTML]{BBBBBB}\textbf{HigherHRNet}} & \multicolumn{4}{c}{\cellcolor[HTML]{BBBBBB}\textbf{ED-Pose}} \\ \midrule 
    \rowcolor[HTML]{D7D7D7} 
    \multicolumn{3}{c|}{\cellcolor[HTML]{D7D7D7}\textbf{Setting}}    & \textbf{val}  & \textbf{val}\tnote{\ddag}  & \textbf{val} \tnote{*\ddag}\quad \! \quad   & \textbf{test} & \textbf{test} \tnote{\ddag}  & \textbf{test} \tnote{*\ddag}\quad \! \quad & \textbf{val}& \textbf{val} \tnote{\ddag}        & \textbf{test}        & \textbf{test} \tnote{\ddag}        & \textbf{val}  & \textbf{test} & \textbf{val}       & \textbf{val} \tnote{*}       & \textbf{test}   & \textbf{test} \tnote{*}       \\ \midrule
    \rowcolor[HTML]{EFEFEF} 
    \multicolumn{3}{c|}{\cellcolor[HTML]{EFEFEF}MSCOCO~\cite{lin2014microsoft}}     &  \textbf{75.6} & 77.6        & 77.2    &  \textbf{73.4} & -    & -     &  79.8        & 82.3       &  81.1        & - & 68.6 & 70.5 & 71.6  & 72.4    & 69.8  & -        \\
    \cellcolor{LightGreen}    & \cellcolor{LightYellow}    & Cartoon       & 9.7 & 37.6& 64.7  & 7.3 & 34.4 & 61.0   & 16.3        & 55.1       & 13.0        & 50.5       & 15.7 & 12.0  & 22.2   & 60.4       & 18.0   & 57.1      \\
    \cellcolor{LightGreen}    & \cellcolor{LightYellow}    & Digital Art   & 22.6 & 59.6& 74.4  & 25.8 & 61.2 & 75.7   & 29.0        & 69.9       & 31.9        & 72.2       & 42.5 & 44.3 & 43.5    & 71.4      & 45.6   & 75.1       \\
    \cellcolor{LightGreen}    & \cellcolor{LightYellow}    & Ink Painting  & 6.3 & 51.4& 72.1 & 5.6 & 48.0 & 72.4   & 8.9        & 59.8       & 9.2        & 58.2       & 26.8 & 20.9 & 28.2  & 56.8  & 24.9   & 55.4  \\
    \cellcolor{LightGreen}    & \cellcolor{LightYellow}    & Kids Drawing  & 10.5 & 40.8& 86.1  & 10.0 & 44.6 & 85.9   & 14.0        & 59.2       & 13.2       & 62.6       & 12.6 & 13.8 & 20.7    & 76.7      & 23.2   & 78.8      \\
    \cellcolor{LightGreen}    & \cellcolor{LightYellow}    & Mural         & 11.6 & 54.0& 71.1  & 12.3 & 53.6 & 71.6   & 15.9        & 50.1       & 16.4      & 51.6       & 30.6 & 32.0 & 34.6   & 64.7      & 35.1   & 65.4       \\
    \cellcolor{LightGreen}    & \cellcolor{LightYellow}    & Oil Painting  & 31.6 & 65.7& 78.1  & 28.5 & 62.2 & 75.6   & 39.6       & 73.4       & 36.7        & 70.5       & 54.4 & 51.1 & 56.2   & 75.2      & 51.7   & 71.4      \\
    \cellcolor{LightGreen}    & \cellcolor{LightYellow}    & Shadow Play   & \underline{ 5.4}  &  \underline{15.9}&  \underline{59.8}  & \underline{5.0}  &  \underline{17.2} &  \underline{58.4}   &  \underline{8.1}&  \underline{29.2}       &  \underline{8.4}        & \underline{ 26.0}       &  \underline{4.4 } & 6.5 &  \underline{6.0}    & \underline{38.5}       &  \underline{7.7}    & \underline{39.7}     \\
    \cellcolor{LightGreen}    & \cellcolor{LightYellow}    & Sketch        & 6.3  & 44.1& 73.1  &  6.7 & 57.2 & 79.4   & 9.1       & 61.3       & 10.9        & 71.3       & 13.6  &  \underline{6.3}  & 12.0     & 66.8      & 12.2  & 75.8     \\
    \cellcolor{LightGreen}    & \cellcolor{LightYellow}    & Stained Glass & 10.4 & 46.0& 74.8  & 9.7 & 45.1 & 73.1   & 12.0        & 59.1       & 12.1        & 58.1       & 26.6 & 23.1 &27.6  & 74.4      & 25.6  & 71.5      \\
    \cellcolor{LightGreen}    & \cellcolor{LightYellow}    & Ukiyoe        & 17.8 & 48.1& 82.4  & 18.8 & 47.7 & 81.8   & 23.8       & 61.2       & 26.8        & 63.1       & 20.2  & 19.4  & 25.0  & 83.6       & 25.8 & 83.6     \\ 
    \cellcolor{LightGreen}    & \multirow{-11}{*}{\cellcolor{LightYellow}\rotatebox{90}{2D Representation}} & Watercolor    & 26.7 & 60.1& 73.9  & 25.5 & 57.6 & 73.4   & 36.4        & 71.0       & 36.1       & 69.0       & 48.9 & 43.4 & 50.6    & 73.5      & 45.6  & 71.3    \\
    \cellcolor{LightGreen}    & \cellcolor{LightBlue}    & Garage Kits   & 45.2 & 57.5& 86.7  & 44.5 & 61.4 & \textbf{89.2}   & 52.5        & 76.2       & 50.6        & 77.0       & 37.4 & 34.7 & 47.9   & 87.7      & 44.1   & 90.1      \\
    \cellcolor{LightGreen}    & \cellcolor{LightBlue}    & Relief        & 10.5 & 57.3& 78.7  & 7.9 & 53.4 & 76.0   & 16.2        & 70.8       & 14.9        & 67.1       & 32.5 & 29.8 & 28.0   & 70.6       & 27.1   & 67.6       \\
    \multirow{-14}{*}{\cellcolor{LightGreen}\rotatebox{90}{Artificial Scene}} & \cellcolor{LightBlue}    & Sculpture     & 36.4 & 65.9& 81.0  & 38.5 & 64.0& 78.5   & 34.9        & 78.5       & 34.2       & 73.7       & 33.5 & 35.2 & 45.9   & 76.9       & 46.7    & 74.7     \\
    \cellcolor{LightRed}    & \cellcolor{LightBlue}    & Acrobatics    & 45.8 & 68.0& 85.2& 46.6 & 68.4& 83.2   & 69.1        & 86.8       & 66.3       & 83.9       & 58.6 & 57.4 & 41.4   & 80.0      & 44.4   & 78.9       \\
    \cellcolor{LightRed}    & \cellcolor{LightBlue}    & Cosplay       & 71.0 &  \textbf{81.1}&  \textbf{87.2}  & 72.6 &  \textbf{81.9}&  87.0   & \textbf{80.0}        & \textbf{90.3}       & \textbf{81.7}        &  \textbf{88.8}       &  \textbf{78.1} &  \textbf{77.8} &  \textbf{79.6}   & \textbf{89.1}       &  \textbf{79.7}    & \textbf{90.4}      \\
    \cellcolor{LightRed}    & \cellcolor{LightBlue}    & Dance& 43.1 & 67.3& 77.2 & 49.2 & 70.1 & 80.1   & 57.3        &  81.5       & 61.5       & 83.8       & 51.4 & 62.4 & 53.6   & 76.5      & 61.2  & 82.2        \\
    \cellcolor{LightRed}    & \cellcolor{LightBlue}    & Drama& 45.3 & 75.1& 82.0  & 46.7 & 75.8& 83.1  & 54.2        & 83.9       & 56.9     & 84.8       & 69.6 & 72.2 & 75.0   & 85.9      & 76.0  & 86.1      \\
    \multirow{-5}{*}{\cellcolor{LightRed}\rotatebox{90}{Natural Scene}}        & \multirow{-8}{*}{\cellcolor{LightBlue}\rotatebox{90}{3D Representation}}  & Movie& 49.5 & 71.5& 77.2 & 50.4 & 72.2& 76.2   & 57.6        & 76.8       & 56.5        & 78.6       & 64.9 & 65.8 & 69.2   & 82.2     & 68.2   & 80.4      \\ \midrule
    \multicolumn{3}{c|}{Average}    & 22.2 & 55.2& 76.4  & 24.1 & 55.4 & 76.0   & 28.7       & 67.7       & 30.7       & 67.5       & 34.6 & 36.3 & 37.5   & 72.3      & 39.2    & 72.7     \\ \bottomrule
    \end{tabular}
    \begin{tablenotes}
   \footnotesize
   \item[\ddag] the top-down pose estimation results that use ground truth bounding box;
   \item[*] the baseline results we provide by training on the joint of MSCOCO~\cite{lin2014microsoft} and \DatasetName. 
    \end{tablenotes}
    \end{threeparttable}}
    
    \caption{Comparisons of average precision (AP) of wildly used human pose estimation models, including top-down methods HRNet and ViTPose~\cite{sun2019deep,xu2022vitpose}, bottom-up method HigherHRNet~\cite{cheng2020higherhrnet}, and one-stage method ED-Pose~\cite{yang2023explicit}. The best results are shown in \textbf{bold} and the worst results are highlighted with \underline{underlined font}. Detailed settings for each model are provided in supplementary files.}
    \label{tab:pose_estimation}
    
 \vspace{-14pt}
    \end{table*}

    We conduct comprehensive experiments on \DatasetName~with several popular human-centric tasks. In each subsection, we introduce the task and related methods first. Then, we present the corresponding results and analyses.

    \subsection{Human-Centric Recognition}

    \subsubsection{Human Detection}
    \label{sec:experiments_object_detection}

    Human detection task~\cite{lin2014microsoft,bodyhands_2022} identifies the bounding box of each person in a given image, which is
    fundamental for further human scene understanding. It is also a crucial step for downstream tasks such as top-down human pose estimation~\cite{Xiao_2018_simple,xu2022vitpose,zeng2022deciwatch}.
    Most object detectors (e.g., YOLO~\cite{redmon2016you}, DETR~\cite{carion2020detr}, and DINO~\cite{zhang2022dino}) do not differentiate humans from other objects in detection. 
    Recently, HBA~\cite{bodyhands_2022} is specifically designed for human and hand detection.

    Tab.~\ref{tab:human_detection} shows the performance of both CNN-based and Transformer-based widely-used detectors on the validation and test sets of \DatasetName. All the pre-trained models have poor performance on artificial scenes, with average precision (AP) ranging from $11.7$\% to $14.7$\%, confirming the impact of the domain gap on the models' generalization ability. 
    For some natural scenarios with similar distribution to the MSCOCO dataset~\cite{lin2014microsoft}, e.g., Dance and Acrobatics, existing models achieve satisfactory performance.
    In particular, the stage backdrop in Acrobatics is usually clean, resulting in a higher AP value compared to that of other categories. At the same time, despite Shadow Play also having a spotless background, its performance is among the lowest because of the huge texture disparity from the natural scene.

 \begin{figure*}[htbp]
    \centering
    \includegraphics[width=0.99\linewidth]{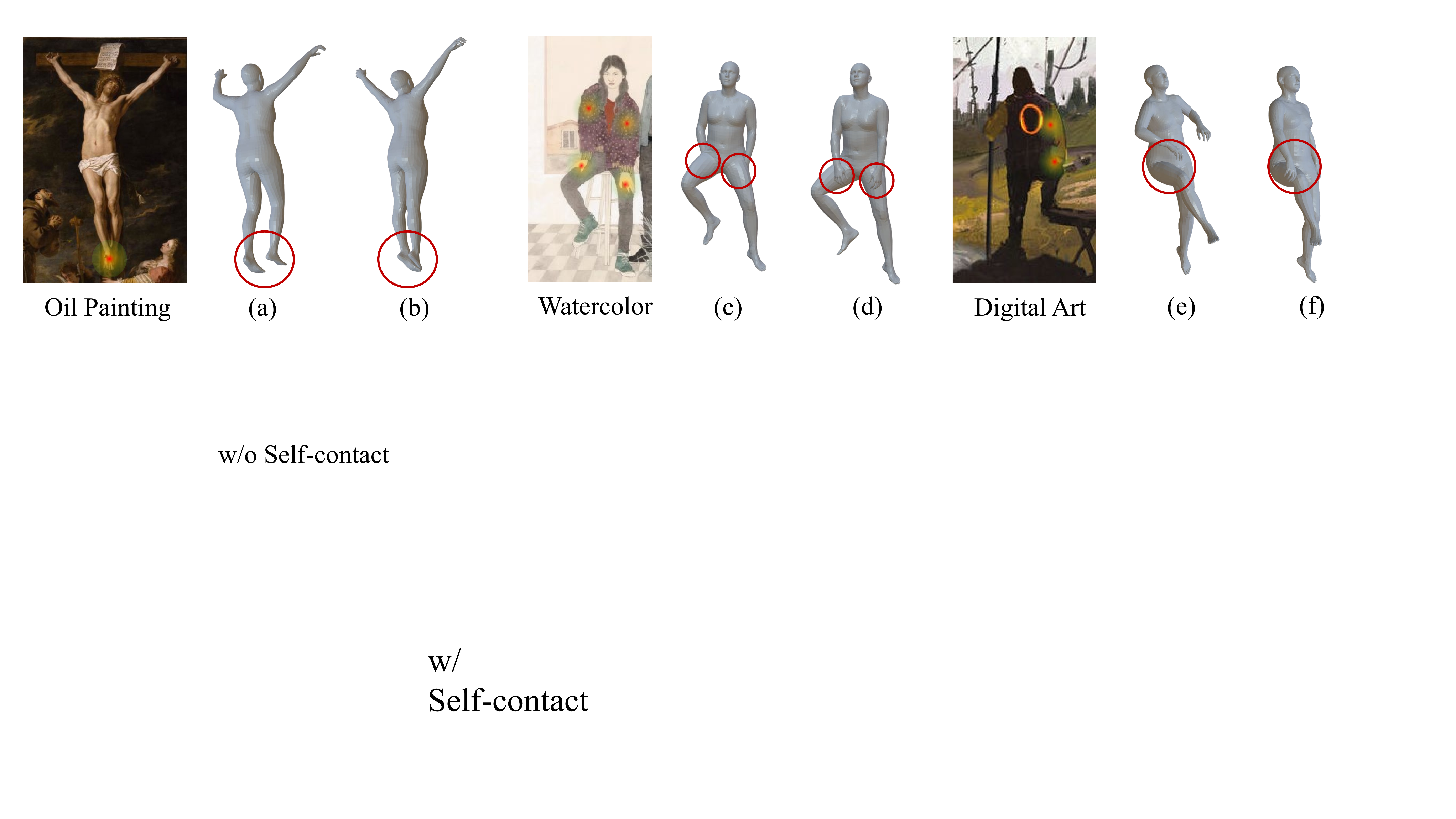}
    \vspace{-5pt}
    \caption{Illustration of how the annotated self-contact points can benefit 3D human mesh recovery. (a), (c), and (e) show the human mesh outputs from three scenes without self-contact optimization. (b), (d), and (f) are optimized mesh results with self-contact points.}
    \label{fig:mesh_recovery}

    \end{figure*}

  \begin{figure*}[h]
        \centering
         \includegraphics[width=1.\linewidth]{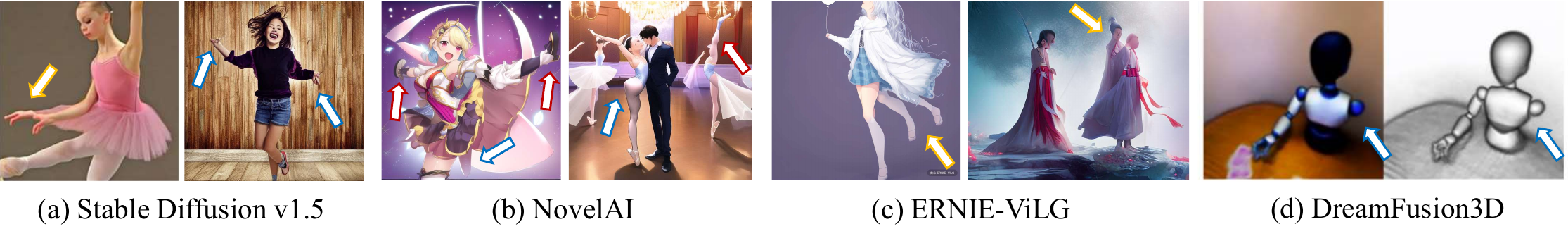}
         \vspace{-0.2cm}
        \caption{Failure cases of existing popular text-to-image diffusion models on human-centric generation. We highlight these cases by \textbf{{\color{Blue}blue}}, \textbf{{\color{Yellow}yellow}}, and \textbf{{\color{Red}red}} arrows for the \textbf{{\color{Blue}missing}}, \textbf{{\color{Yellow}redundant}}, and \textbf{{\color{Red}replaced}} body parts. We simply regard the natural human structure as more desirable, despite the fact that there is no right or wrong in art.}
        \label{fig:exp_fail}
        \vspace{-0.1cm}
        \end{figure*}

    Moreover, we provide a baseline model by training Faster R-CNN on the joint of MSCOCO~\cite{lin2014microsoft} and \DatasetName. The joint training procedure leads to about a 56\% performance boost in Shadow Play and a 31\% average improvement in all categories.  
    Nevertheless, the performance of the baseline model on \DatasetName~is still relatively low, calling for future research on this topic.

    \subsubsection{2D Human Pose Estimation}
    \label{sec:experiments_2d_pose_estimation}

    Human Pose Estimation (HPE) is another basic task for human motion analysis, which can be divided into 2D HPE and 3D HPE, outputting 2D keypoints and 3D keypoints respectively. 
    Hard poses, heavy occlusions, and confusing backgrounds make these tasks still quite challenging after years of research. 
    Existing 2D HPE methods can be categorized into three types: top-down~\cite{newell2016stacked,sun2019deep,xu2022vitpose}, bottom-up~\cite{cheng2020higherhrnet,cao2017realtime}, and one-stage~\cite{yang2023explicit}.
    Generally speaking, top-down approaches~\cite{newell2016stacked,sun2019deep,xu2022vitpose} usually have higher performance than other methods, provided that human detection performs correctly. However, they suffer from high computational costs. 
    In contrast, bottom-up methods~\cite{cheng2020higherhrnet,cao2017realtime} are efficient, especially for crowded scenes, but have relatively low accuracy.
    To trade off efficiency and effectiveness, one-stage methods (e.g. PETR~\cite{shi2022end}, ED-Pose~\cite{yang2023explicit}) are proposed thanks to the emergent DETR-based models~\cite{Zhu_detr21}. 
    We provide results for these representative methods in Tab.~\ref{tab:pose_estimation}.

\begin{figure}[H]
        \vspace{-2pt}
            \centering
             \includegraphics[width=1.\linewidth]{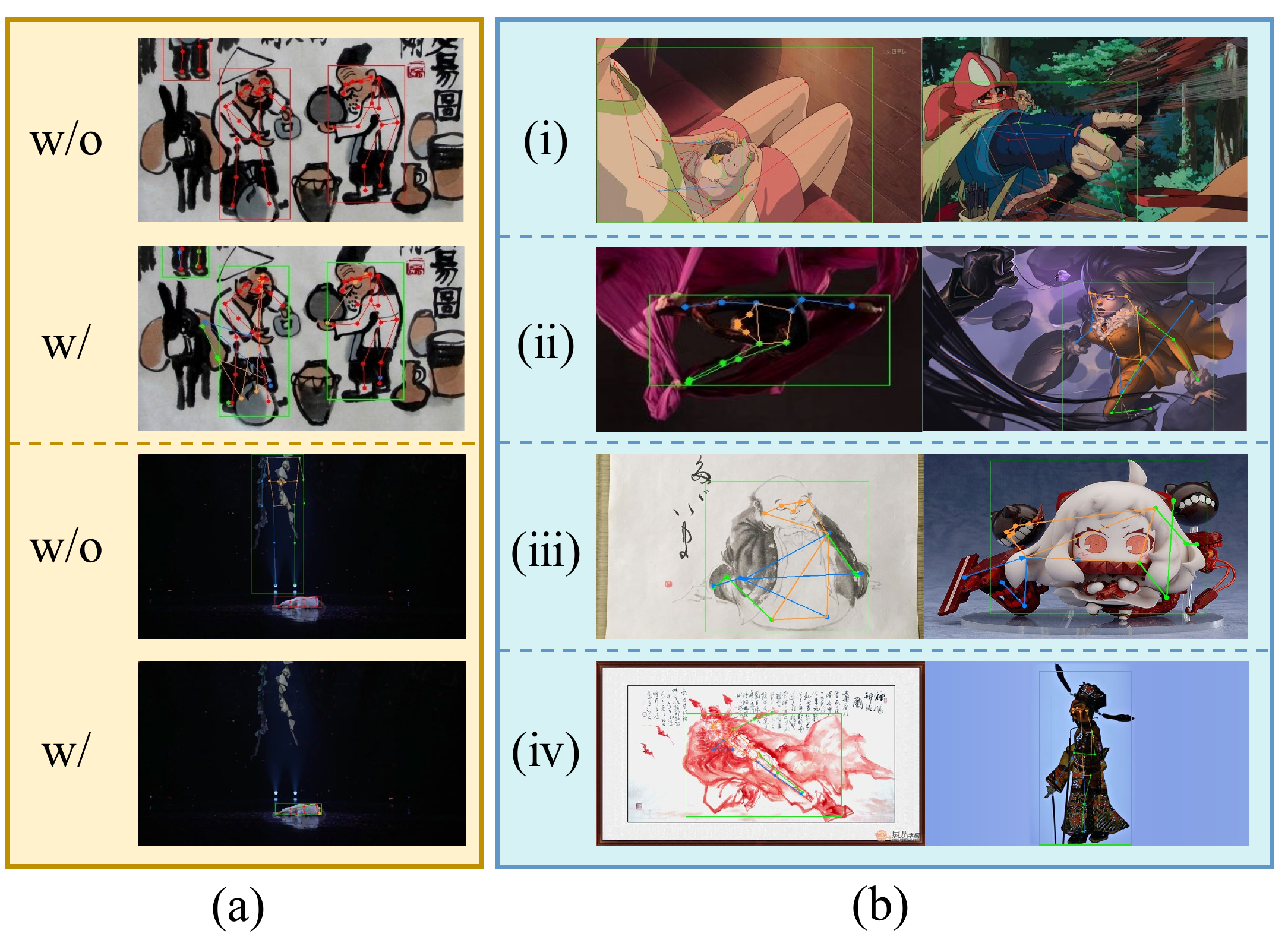}
            \caption{Failure cases of the pose estimator HRNet on \DatasetName. The first column presents how human detection impacts top-down pose estimation. Red lines and points represent ground truth, lines and points in other color are detected results. \emph{w/o} shows pose estimation results based on detected boxes, \emph{w} means results with the ground truth bounding box. The right figures show   (\romannumeral1) perspective, (\romannumeral2) pose, (\romannumeral3) shape, and (\romannumeral4) texture in \DatasetName~ are challenging to existing pose estimators (trained on MSCOCO).}
                 \vspace{-0.5cm}
            \label{fig:detected_results}
            \end{figure}

    Specifically, we show the quantitative results for widely used as well as the SOTA pose estimation methods on the validation and testing sets of \DatasetName. Top-down human estimation depends heavily on the accuracy of human detection, leading to performance elevation when the ground-truth bounding box is given, as shown in Fig.~\ref{fig:detected_results} (a). Different from the human detection task, pose complexity has a bigger impact on results than the image background. Although cosplay typically includes a complex image background, simple postures ease the estimation procedure and lead to high estimation accuracy. Shadow Play still shows a low estimation accuracy due to the large shape and texture differences from humans in natural scenes. Some pose failure cases are shown in Fig.~\ref{fig:detected_results} (b).

    Moreover, we provide a baseline model by training HRNet on the joint of MSCOCO and \DatasetName, resulting in an overall 21\% boost in accuracy. Notably, after training with ED-Pose, results on MSCOCO raise 0.8, indicating multi-scenario images may benefit feature extraction and human understanding of real scenes.

    \subsubsection{Human Mesh Recovery}
    \label{sec:experiments_3d_pose_estimation}
\vspace{-2pt}

    Statistical body models such as SMPL~\cite{SMPL:2015} show their convenient usage for animation, games, and VR applications. 
    These models represent humans in a watertight and animatable 3D human body mesh with a small number of parameters, which largely simplifies 3D human mesh expression.
    However, depth ambiguities hinder the fidelity of 3D human mesh estimation from a monocular camera.
    To overcome this issue, similar to Sketch2Pose~\cite{brodt2022sketch2pose}, we further provide self-contact annotations as additional information to facilitate reasonable depth optimization via the interpenetration penalty.
    By mapping the contact region onto the vertices of a rough SMPL model generated by Exemplar Fine-Tuning (EFT)~\cite{DBLP:journals/corr/abs-2004-03686} and then minimizing the distance among the contact vertices, visualization results in Fig.~\ref{fig:mesh_recovery} show how annotating self-contact keypoints benefit 3D mesh recovery.

\subsection{Image Generation}
\label{sec:experiments_generation}

\vspace{-3pt}

Image generation has experienced great advances in the past few years with eye-catching generative vision and language models such as unCLIP \cite{dalle2}, Latent Diffusion \cite{latentdiffusion} and Imagen \cite{imagen}. Such progress is enabled by recent breakthroughs of generative models, as well as large-scale web-crawled datasets such as LAION \cite{laion}. However, despite their impressive capability of generating photorealistic and creative images, they often fail at faithfully respecting human structures, as shown in Fig.~\ref{fig:exp_fail}.

\begin{figure}[H]
    \centering
    \includegraphics[width=1\linewidth]{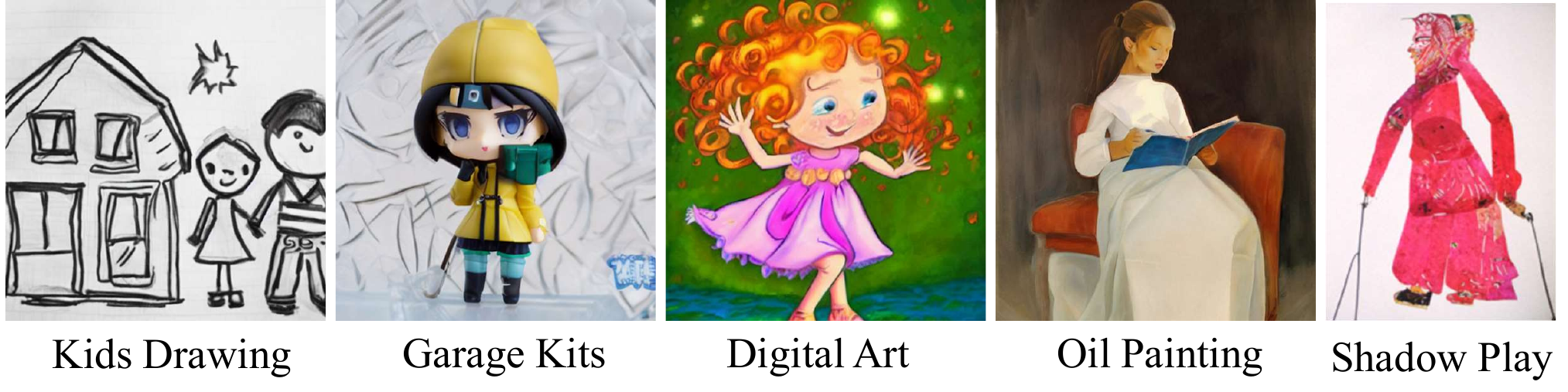}
    \vspace{-0.4cm}
    \caption{Example generations with five scenes from a diffusion generative model trained on \DatasetName. Notably, Shadow Play is a novel scene for existing generative models. }
    \vspace{-0.4cm}
    \label{fig:generation_collage}
    
    \end{figure}

By offering fine-annotated artificial scenarios with \DatasetName, we not only introduce a valuable supplement to existing datasets. More importantly, it provides a good prior for generating art images with plausible human poses. We show some examples produced by fine-tuning a diffusion model on \DatasetName~in Fig.~\ref{fig:generation_collage}. Note that \DatasetName~augments the art category Shadow Play, which is absent from SOTA generative models such as Stable Diffusion.

\subsection{Motion Transfer}
\label{sec:experiments_motion_transfer}

    The motion transfer task aims to generate a new image or video of the source person by learning motion from target images while preserving the source character's appearance. 
    
    Previous motion transfer methods can be roughly divided into two categories. Model-based methods~\cite{ma2017pose,chan2019everybody,sarkar2020neural} use off-the-shelf pose estimators to extract pose information and then use pose skeletons to drive the character. In contrast, model-free methods~\cite{siarohin2019animating,siarohin2019first,tao2022structure} can automatically detect character-agnostic 
    implicit keypoint trajectories to transfer for arbitrary objects.

    \begin{figure}[H]
    \centering
    \includegraphics[width=1.\linewidth]{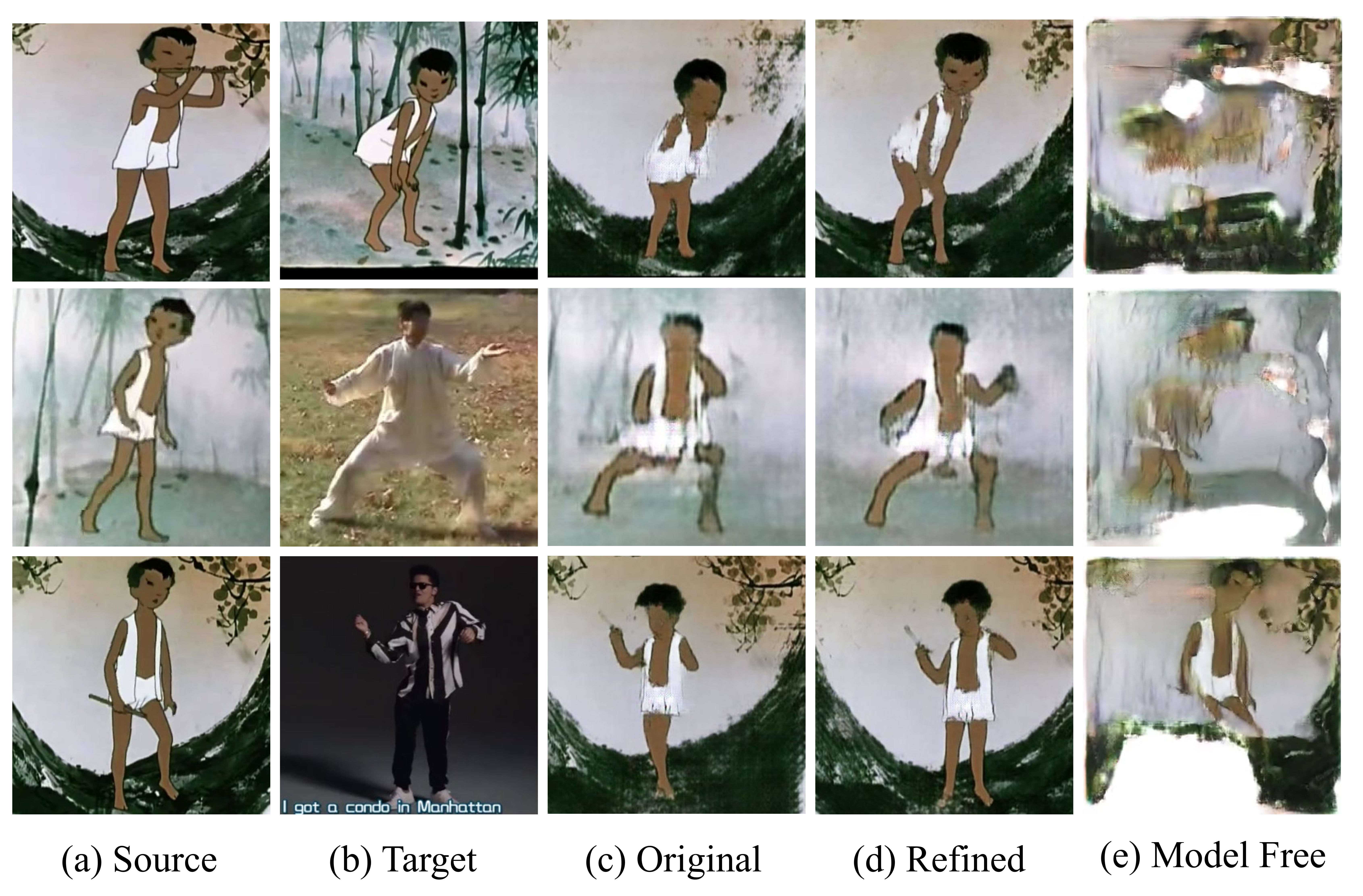}
    \caption{Visualization of model-based and model-free motion transfer results of intra-scene transfer (the 1st row) and inter-scene motion transfer (the 2nd and 3rd rows). The model-free method FOMM~\cite{siarohin2019first} severely fails due to unstable correspondence. In contrast, the model-based methods~\cite{chan2019everybody} are more suitable for multi-scene motion transfer. After using the pose estimation model pre-trained on \DatasetName, (d) shows better results than (c),}
    \label{fig:motion_transfer}
    \vspace{-0.38cm}
    
    \end{figure}
    
    As shown in Fig.~\ref{fig:motion_transfer}~(e), model-free approaches~\cite{siarohin2019first} easily fail on new scenes because of the unstable correspondence between source and driving images. 
    Model-based methods show more stable performance on human motion transfer, but they highly rely on accurate pose estimation results. Present pose estimators are unsuitable for artificial scenes such as kids' drawings\footnote{\url{https://sketch.metademolab.com/}}, thus requiring training a pose estimation model specific to the scene. To illustrate how pose estimators with multi-scenario adaptability can benefit motion transfer tasks, we conduct experiments on the well-known EveryBodyDanceNow~\cite{chan2019everybody} without face enhancement. Although there are model-based models that generate better results than EveryBodyDanceNow, we choose this model to illustrate how poses influence motion transfer because it is widely accepted in the literature. Fig.~\ref{fig:motion_transfer}~(c) shows the original motion transfer result with pose estimator OpenPose\cite{8765346} trained on natural human scenes. By refining the pose estimator with \DatasetName, Fig.~\ref{fig:motion_transfer}~(d) shows how a better pose detection model greatly benefits the motion transfer task.

\section{Conclusion and Discussions}
\label{sec:conclusion_and_discussions}
    \vspace{-3pt}

In this paper, we have presented \DatasetName, a rich-scenario human-centric dataset containing $50k$ high-quality images with versatile manual annotations, which serves as a new challenging dataset for multiple computer vision tasks, such as human detection, human pose estimation, body mesh recovery, motion transfer, and image generation. 
In our experiments, we provide comprehensive baseline results and detailed analyses for these tasks.
We hope that this work will shed light on related research areas and open up new research questions.

\vspace{5pt}
\textbf{Limitations and Future work:} Images in \DatasetName~could be misused for generating fake images, which may bring negative social impact. Moreover, although downstream tasks have been extensively explored on \DatasetName, we simply conduct experiments to reveal how and why existing methods often fail on our dataset, but did not offer a superior solution. Therefore, there is a significant gap to fill for these rich-scenario human-centric tasks, calling for novel solutions in future research. Specifically, future directions with \DatasetName~include but are not limited to 1) cross-domain human recognition algorithms that can adapt to different scenes with various human poses, shapes, textures, and image backgrounds; 2) trustworthy image generation with reasonable human body structure, especially controllable human image generation such as GLIGEN~\cite{li2023gligen}and ControlNet~\cite{zhang2023adding}; 3) Inclusive motion transfer algorithms across different scenes.

We plan to continuously expand \DatasetName~to support new scenarios. To facilitate future human-centric studies, we will make the training and validation set public with an easy-to-use data visualization platform. For the test set, we will provide a testing interface but withhold the data to prevent test information leakage.

\vspace{-0.2cm}

\section{Acknowledgements}

\vspace{-0.1cm}

This work was supported in part by the Shenzhen-Hong Kong-Macau Science and Technology Program (Category C) of the Shenzhen Science Technology and Innovation Commission under Grant No. SGDX2020110309500101 and in part by Research Matching Grant CSE-7-2022.


\clearpage


\appendix

\section*{Supplementary Materials}
\noindent This supplementary material presents more details and additional results not included in the main paper due to page limitation. The list of items included are:

\begin{itemize}
    \item More dataset statistics and analysis in Sec.~\ref{sec_data_statistic_and_analysis}, including statistics and analysis of image source, keypoint visible/occlude/invisible attributes distribution, human size distribution, and annotation visualization.
    \item Experimental details in Sec.~\ref{sec:experimental_details}, including implementation details of models used in our experiments, analysis of human detection and pose estimation results, and more evaluation results on \DatasetName.
    \item More discussion about related datasets for multi-scenario generalization in Sec.~\ref{sec:related_work_multi_scenario_datasets}. 
\end{itemize}

\section{Dataset Statistic and Analysis}
\label{sec_data_statistic_and_analysis}
\subsection{Image Sources}

\begin{figure}[h]
    \centering
    \includegraphics[width=0.92\linewidth]{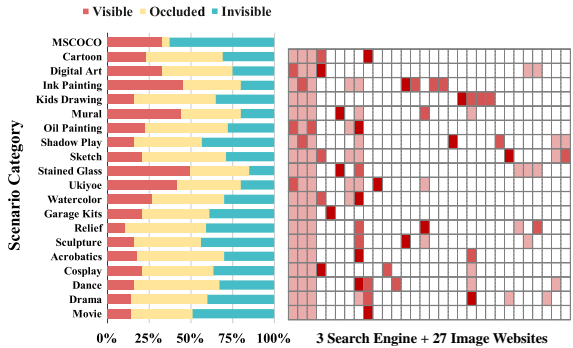}
    \caption{Statistical analyses on the visibility for all keypoints comparing our $20$ scenes with MSCOCO~\cite{lin2014microsoft} (left) and the distribution map of image sources of \DatasetName~(right).}
    \label{fig:scenario_statistic}
    \vspace{-0.8cm}
    \end{figure}

    \DatasetName~is a comprehensive human-centric dataset with 50,000 images from 20 distinctive scenarios, where each scenario contains 2,500 images. As shown in the right part of Fig.~\ref{fig:scenario_statistic}, the images are selected from a total of 30 different image sources and we guarantee diversity of image sources for each scenario. Specifically, \DatasetName~include images collected from European, North American, East Asia, and South African authors, ranging from Before the Common Era to the 21st century with humans in different poses, shapes, and textures.

\subsection{Keypoints Attribute}
    \DatasetName~follows MSCOCO~\cite{lin2014microsoft} to annotate keypoints with visible/occlude/invisible attributes. The left part of Fig.~\ref{fig:scenario_statistic} shows the percentage of the total visible, occluded, and invisible keypoints in all the annotated scenarios of \DatasetName~compared to MSCOCO. As can be observed, 
    the invisible keypoints of the MSCOCO dataset reach a percentage of 63.2\%, which is much higher than that of all the categories in \DatasetName. We attribute it to the fact that MSCOCO does not only focus on human-centric scenes, despite the fact that it contains more than 250,000 humans. 
    This resulted in a large percentage of small-scale, incomplete, and fuzzy humans, which can only be annotated using bounding boxes.
    At the same time, the percentage of visible keypoints of many categories in \DatasetName~is slightly low. This is because, on the one hand, artistic natural scenarios usually contain elaborate movements and fabric coverings, which obstruct the human body; on the other hand, images in artificial scenarios may have unclear lines or missing body pieces lost in history. 
    Overall, \DatasetName~has a higher percentage of keypoint annotation than COCO, which can benefit the related tasks with more valid data.

    \subsection{Human Size Distribution}

    \begin{figure}[t]
    \centering
    \includegraphics[width=.8\linewidth]{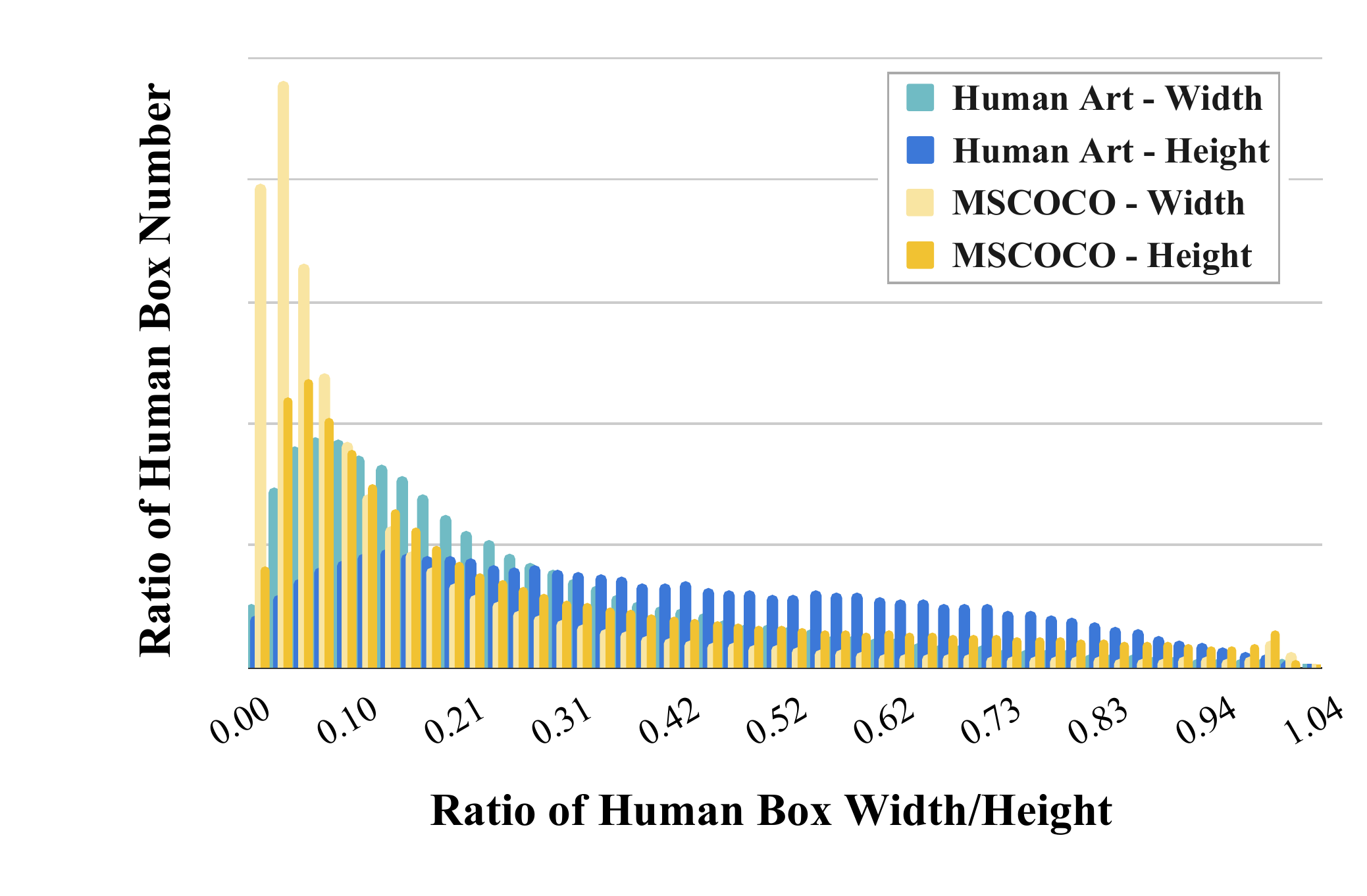}
    \caption{Distribution of ${{bounding\,\,box\,\,width}/{image\,\,width}}$ and ${{bounding\,\,box\,\,height}/{image\,\,height}}$. The horizontal axis shows the ratio of a human bounding box's height and width to the entire image. The vertical axis shows the percentage of human bounding boxes with the corresponding height and width ratio.}
    \label{fig:bbox_size}

    \vspace{-0.5cm}
    
    \end{figure}

    As shown in Fig. \ref{fig:bbox_size}, human sizes in \DatasetName~are more evenly distributed than MSCOCO~\cite{lin2014microsoft}. The average height of humans in \DatasetName~is $0.40$ times the image's height, whereas in MSCOCO is $0.28$ times, which shows that \DatasetName~has fewer tiny humans than MSCOCO due to its human-centric image collecting process. The average ratio for human width is $0.15$ and $0.25$ in \DatasetName~and MSCOCO, respectively. For the reason that human beings are usually long and thin, the width ratio of humans is more concentrated in small proportions. A more balanced distribution of human sizes enables \DatasetName~to support downstream tasks required for various-sized humans. For example, motion transfer usually needs bigger and more detailed human figures to output characters with higher fidelity. However, image generation needs to generate humans in a variety of resolutions to satisfy the user's requirements. More interestingly, 
    despite our relatively large as well as balanced human size, the poor performance from detection and pose estimation suggests that our dataset still offers difficulties beyond scale, such as appearance diversity, background variation, and pose complexity.

    \subsection{Annotation Visualization}

\begin{figure*}[h]
    \centering
    \includegraphics[width=0.95\linewidth]{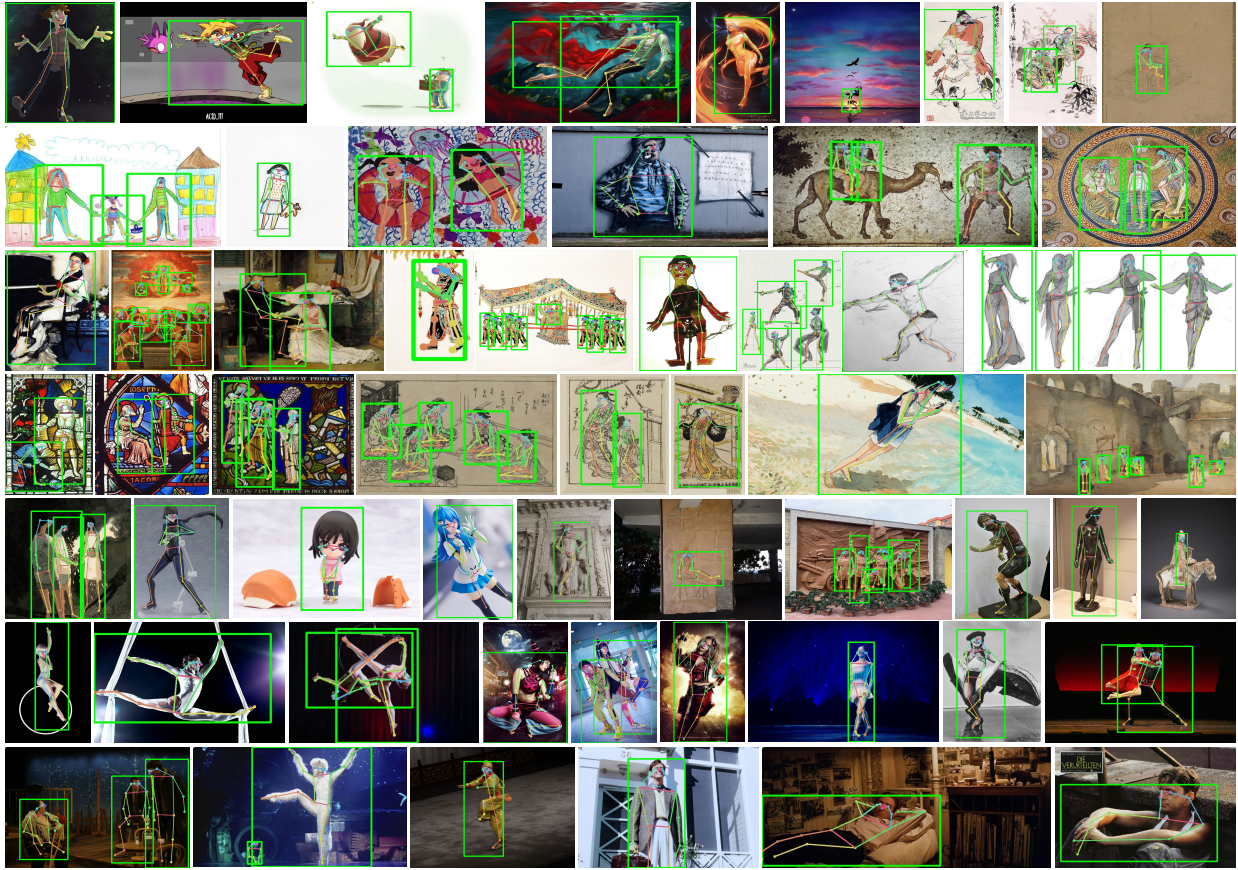}
    \caption{Annotated examples in \DatasetName. We randomly select $3$ images from each category to show the annotation quality and image diversity of \DatasetName. Images in \DatasetName~are varied in terms of human shape, pose, texture and size.}
    \label{fig:annotation}
    \vspace{-0.5cm}
    \end{figure*}

    We show the annotation quality and image diversity of \DatasetName~with the human bounding-box and keypoint annotation in Fig. \ref{fig:annotation}. The diversity of \DatasetName~derives from the wide variations in painting techniques among categories as well as the variations in the human size, body shape, and character pose within each category. This makes \DatasetName~a more challenging dataset than previous real-world human datasets, necessitating higher generalization abilities from the detection and estimation model.

\section{Exprimental Details}
\label{sec:experimental_details}

\subsection{Implementation Details}

For human detection, we provide baselines of Faster R-CNN~\cite{NIPS2015_14bfa6bb}, YOLOX~\cite{redmon2016you}, Deformable DETR~\cite{Zhu_detr21}, and DINO~\cite{zhang2022dino}. All the pretrained models we use are trained exclusively on MSCOCO~\cite{lin2014microsoft}. And the training is implemented on the random shuffle of MSCOCO and \DatasetName.  The implementation details are explained as follows:
\begin{itemize}
    \item Faster R-CNN: We choose Faster R-CNN on Feature Pyramid Networks~\cite{lin2017feature} with ResNet-50~\cite{he2016deep} as the backbone. For testing, the pertained model we use is trained on $8$ NVIDIA GTX 1080 Ti GPUs for $12$ epochs. For training, we trained on $8$ NVIDIA RTX 3090Ti GPUs. Given that data volume has almost doubled, we trained $21$ epochs rather than the original $12$ epochs to guarantee model convergence.
    \item YOLOX: We choose YOLOX-L with an input size of $640$x$640$. For testing, the pertained model we use is trained on $8$ NVIDIA Tesla PG503-216 GPUs for $300$ epochs. For training, we trained on $4$ Nvidia Tesla A100 GPUs.
    \item Deformable DETR: We choose Two-Stage Deformable DETR with ResNet-50~\cite{he2016deep} as the backbone. For testing, the pertained model we use is trained on $8$ Nvidia Tesla v100 GPUs for $50$ epochs. For training, we trained on $4$ Nvidia Tesla A100 GPUs.
    \item DINO: We choose DINO-5scale with Swin-L~\cite{liu2021swin} as the backbone. For testing, the pertained model we use is trained on Nvidia Tesla A100 GPU for $31$ epochs.
    
\end{itemize}

For human pose estimation, we provide baselines of HRNet~\cite{sun2019deep},  ViTPose~\cite{xu2022vitpose}, HigherHRNet~\cite{cheng2020higherhrnet}, and ED-Pose~\cite{yang2023explicit}. All the pretrained models we use are trained exclusively on MSCOCO~\cite{lin2014microsoft}. The training is implemented on the random shuffle of MSCOCO and \DatasetName. For top-down pose estimation methods, we use human detectors with settings the same as listed above in testing, and use augmentations of the ground truth bounding box (e.g., random flip, random bounding box center shift) in the testing and training stage. To make a fair comparison with COCO, testing and training only consider 17 human keypoints. The implementation details are explained as follows:
\begin{itemize}
    \item HRNet: We choose HRNet-W48 with an input size of $256$x$192$. For testing, the pertained model we use is trained on $8$ Nvidia Tesla v100 GPUs for $210$ epochs. For training, we trained on $4$ Nvidia Tesla v100 GPUs.
    \item ViTPose: We choose ViTPose-H with an input size of $256$x$192$. For testing, the pertained model we use is trained on $8$ Nvidia Tesla v100 GPUs for $210$ epochs.
    \item HigherHRNet: We choose HigherHRNet-W48 with an input size of $512$x$512$. For testing, the pertained model we use is trained on $8$ Nvidia Tesla v100 GPUs for $300$ epochs. For training, we trained on $4$ Nvidia Tesla A100 GPUs.
    \item ED-Pose: We choose ED-Pose with ResNet-50~\cite{he2016deep} as the backbone. For testing, the pertained model we use is trained for $60$ epochs.
    
\end{itemize}

    \begin{figure*}[htbp]
    \centering

    \begin{minipage}[t]{0.5\linewidth}
    \centering
    \includegraphics[width=0.95\linewidth]{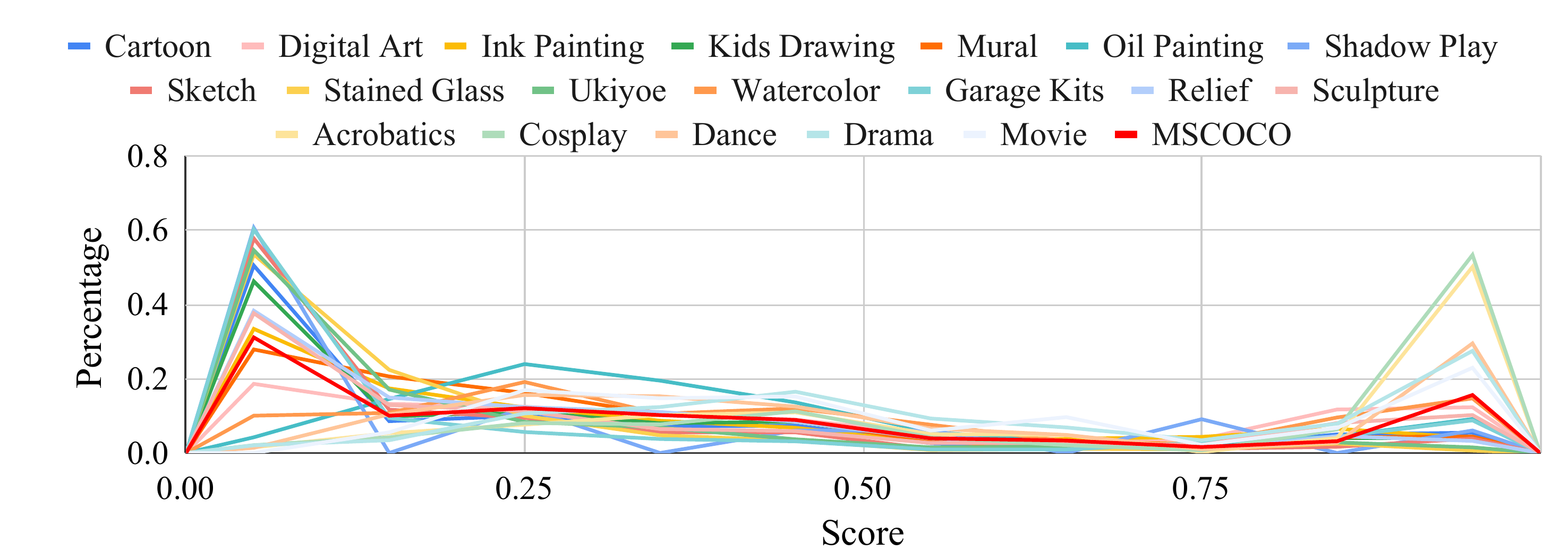}
    \subcaption{Human Detection Confidence Score Distribution before Training}
    \end{minipage}%
    \begin{minipage}[t]{0.5\linewidth}
    \centering
    \includegraphics[width=0.95\linewidth]{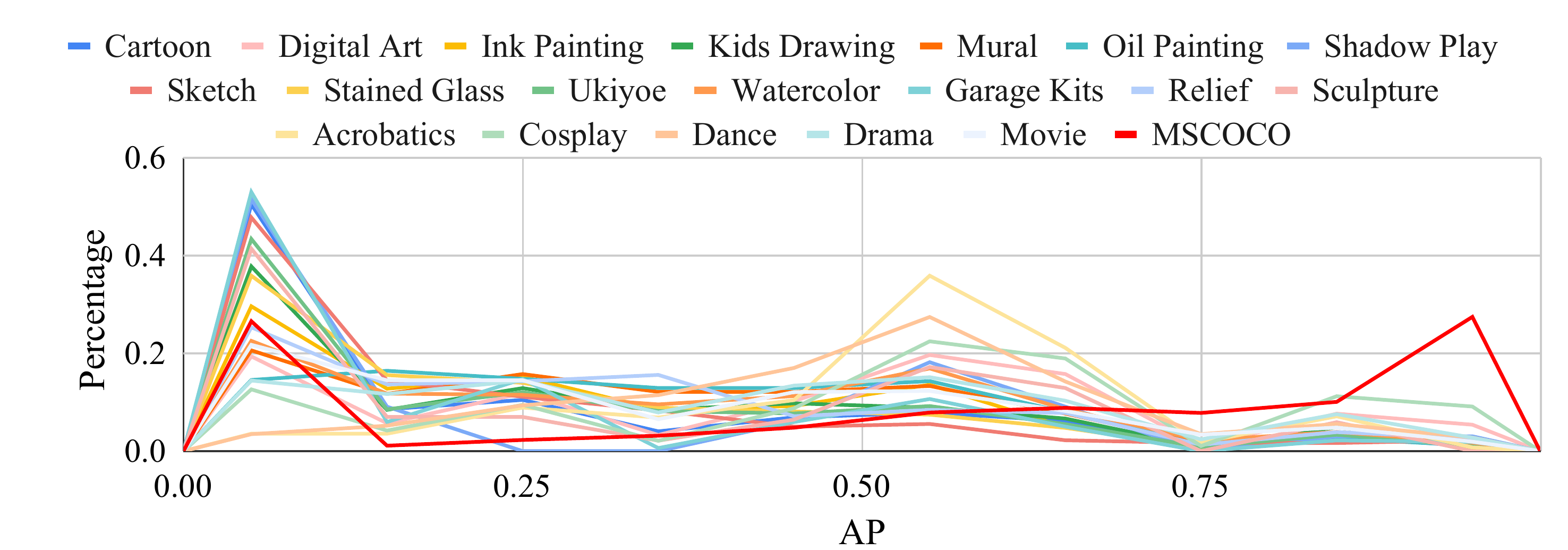}
    \subcaption{Human Detection AP Distribution before Training}
    \end{minipage}%

    \begin{minipage}[t]{0.5\linewidth}
    \centering
    \includegraphics[width=0.95\linewidth]{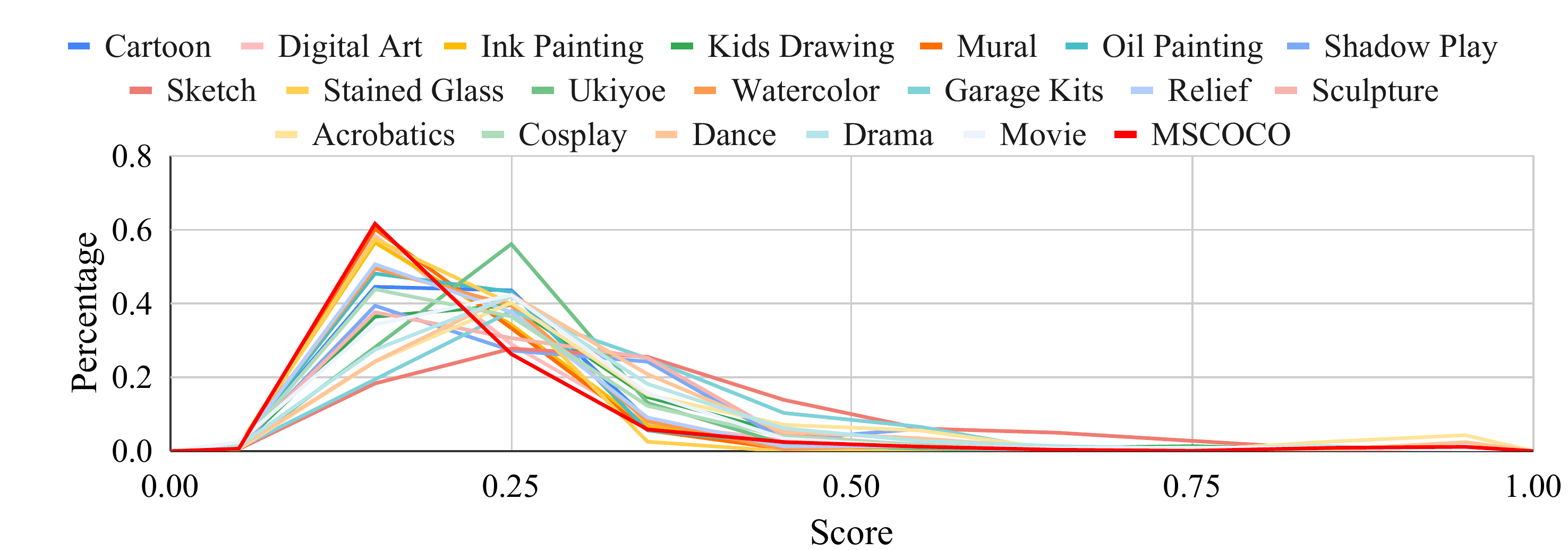}
    \subcaption{Human Detection Confidence Score Distribution after Training}
    \end{minipage}%
    \begin{minipage}[t]{0.5\linewidth}
    \centering
    \includegraphics[width=0.95\linewidth]{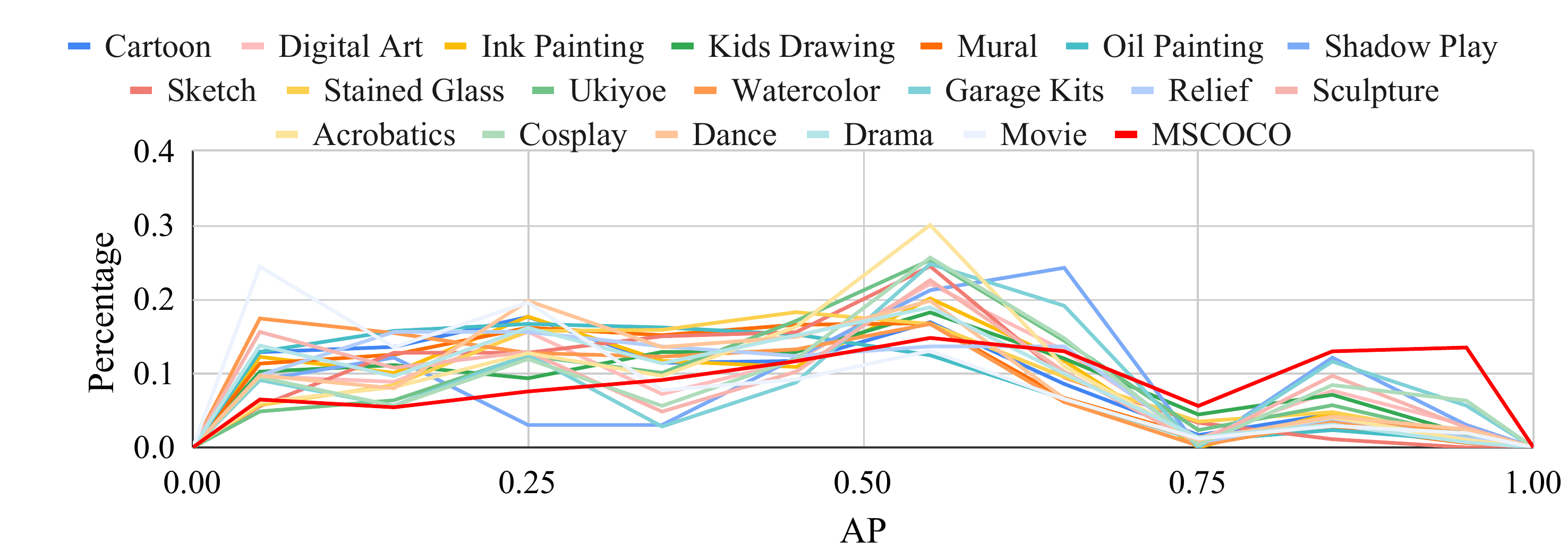}
    \subcaption{Human Detection AP Distribution after Training}
    \end{minipage}%
    
    \begin{minipage}[t]{0.5\linewidth}
    \centering
    \includegraphics[width=0.95\linewidth]{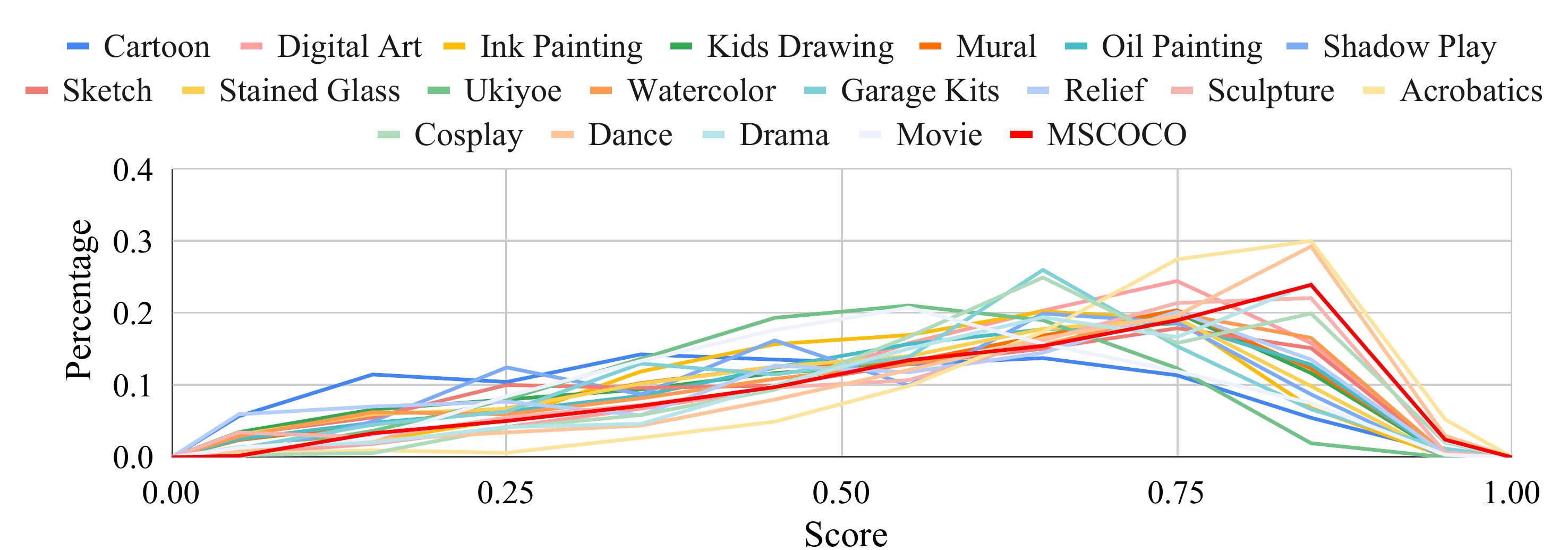}
    \subcaption{Pose Estimation Confidence Score Distribution before Training}
    \end{minipage}%
    \begin{minipage}[t]{0.5\linewidth}
    \centering
    \includegraphics[width=0.95\linewidth]{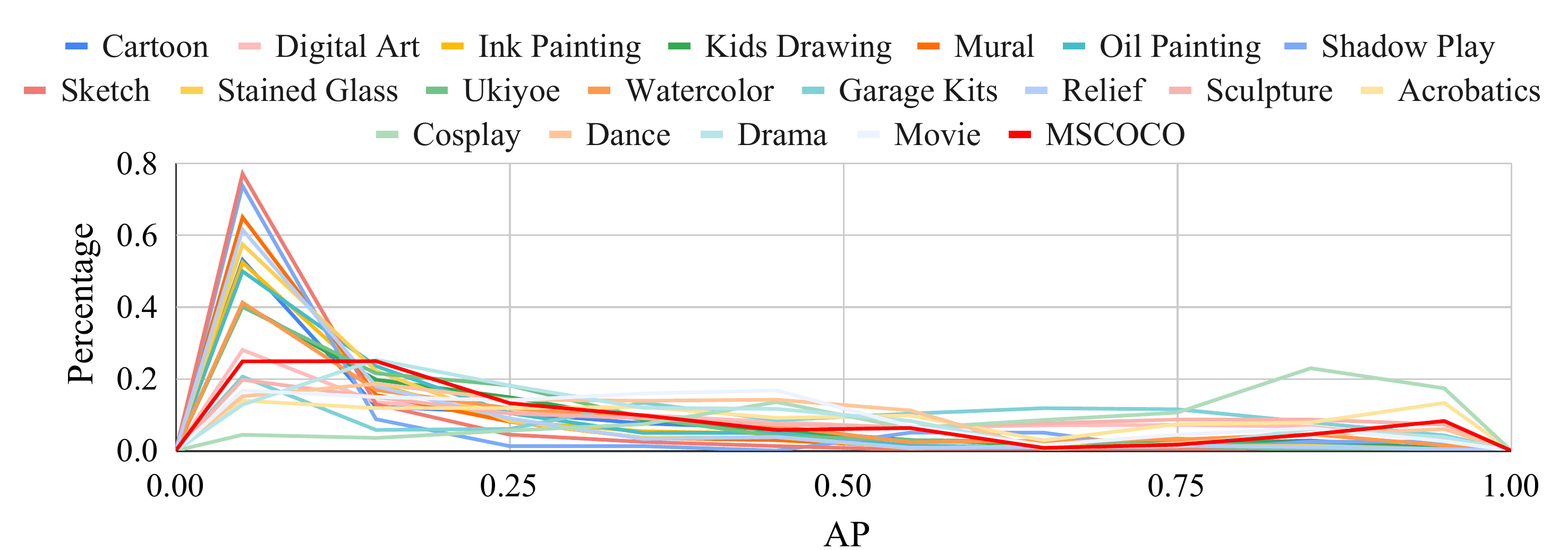}
    \subcaption{Pose Estimation AP Distribution before Training}
    \end{minipage}%

    \begin{minipage}[t]{0.5\linewidth}
    \centering
    \includegraphics[width=0.95\linewidth]{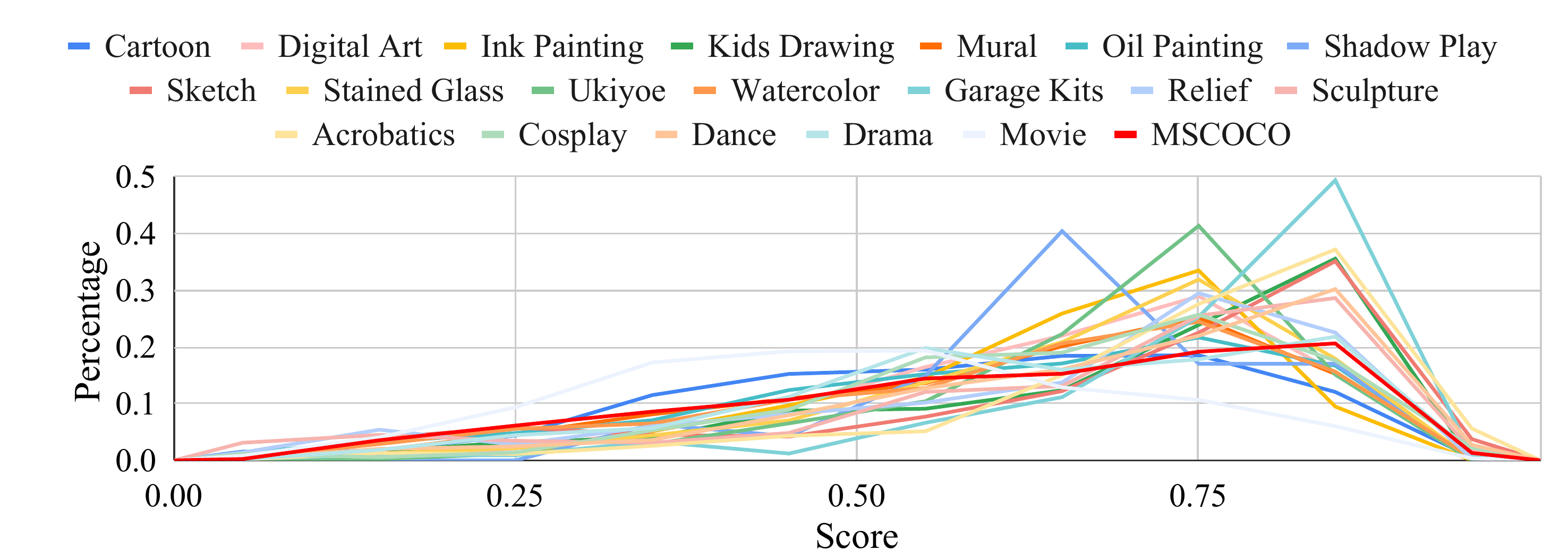}
    \subcaption{Pose Estimation Confidence Score Distribution after Training}
    \end{minipage}%
    \begin{minipage}[t]{0.5\linewidth}
    \centering
    \includegraphics[width=0.95\linewidth]{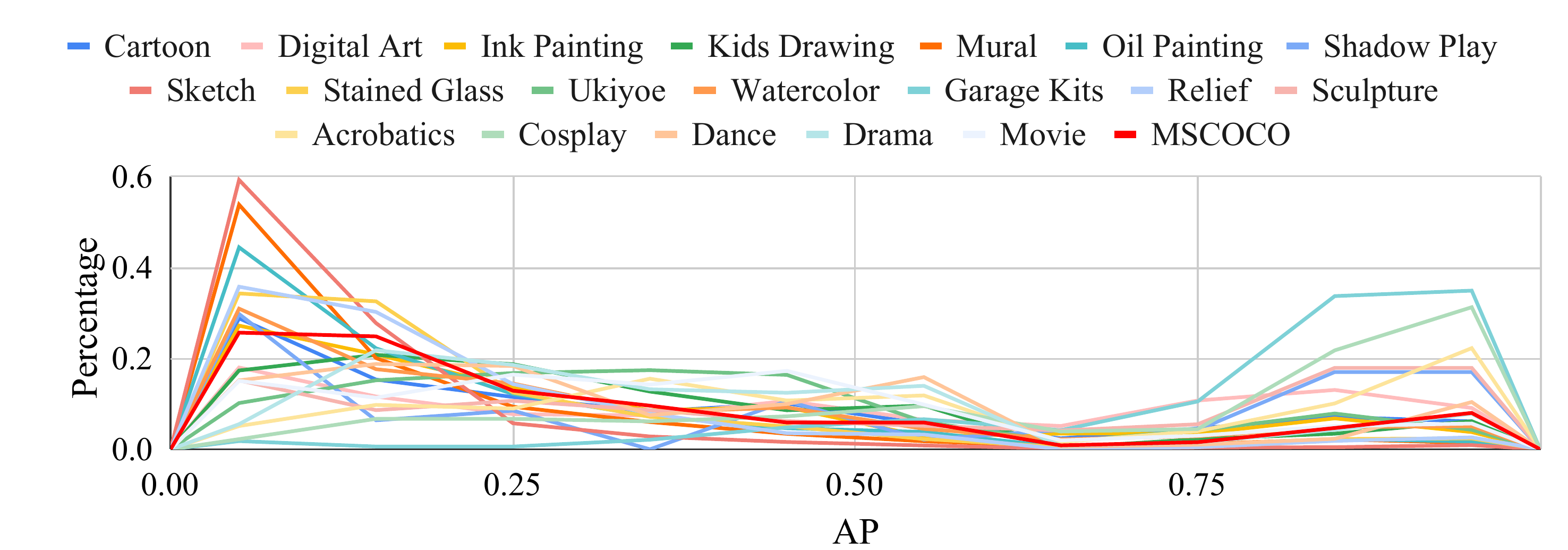}
    \subcaption{Pose Estimation AP Distribution after Training}
    \end{minipage}%
    \centering
    \caption{Contrast of confidence score and AP distribution of human detection model YOLOX~\cite{redmon2016you} and pose estimation model HRNet~\cite{sun2019deep} before and after training on our proposed scenes and COCO. Specifically, (a)-(d) shows the distribution of pose estimation, and (e)-(h) shows the distribution of human detection. The horizontal axis of each figure shows the confidence score/AP intervals. The vertical axis of each figure shows the image percentage in each score/AP interval.}
    \label{fig:ap_score}
    \vspace{-0.5cm}
    \end{figure*}
    
For human mesh recovery, we use the same optimization strategies as in Sketch2Pose~\cite{brodt2022sketch2pose} with $17$ human keypoints and self-contact keypoint. The 2D-to-SMPL model used in Sketch2Pose~\cite{brodt2022sketch2pose} includes optimization of 2D bone tangents, body part contacts, and bone foreshortening. In the main paper, we contrast the visualization results of not using and using body part contacts in Fig. 6. Results show using self-contact keypoints benefits 3D mesh recovery by minimizing the 3D distance near the self-contact area. Noted that due to the influence of the other two optimization elements, bone tangents and bone foreshortening, body parts that do not directly connect to the self-contact area show different poses with and without the self-contact optimization (e.g. the elbow angle in Fig. 6 (a) and (b) in the main paper).

    \subsection{More Analyses of Human-centric Tasks }
    
    The confidence scores output from pose estimation or human detection model indicate how confident the model is for output results. The AP scores indicate the models' average prediction accuracy. We try to analyze why these current models underperform on our data based on the two metrics. Fig.~\ref{fig:ap_score} shows the variation of human detection model YOLOX~\cite{redmon2016you} and pose estimation model HRNet~\cite{sun2019deep}'s confidence score and AP distribution before and after training. We use the same trained model as in main paper.
    
    We provide analyses from the following three aspects: (1) The contrast of confidence score and AP distribution. We discover that the model tends to be over-confident, where the confidence score distribution and AP distribution do not show a positive correlation. This issue has become more serious in artificial scenes of \DatasetName. For instance, in Fig.~\ref{fig:ap_score}~(e) and Fig.~\ref{fig:ap_score}~(f), although the pose estimation model shows a relatively high confidence score in most images, a large proportion of the estimation outputs' AP scores range from $0$ to $0.25$. (2) The contrast between before training and after training. The recurring finding is that training can reduce the percentage of both low confidence scores and low AP scores, which is consistent with common sense. Another interesting finding is that, although the mean AP score on MSCOCO~\cite{lin2014microsoft} is reduced after joint training, the percentage of the low scores is reduced as well, as shown in Fig.~\ref{fig:ap_score}(b) and Fig.~\ref{fig:ap_score}(d). This may be because the more evenly distributed human size and the richer depictions in \DatasetName~help the model to obtain better adaptability on hard poses in real-world scenarios.
    (3) The contrast between human detection and pose estimation tasks. After training, human detection shows a more uniform AP distribution along the horizontal axis. However, the pose estimation model shows concentrated distributions in low and  high AP scores. This may be due to the differences in the two methods' targets. When human detection fails, valid interactions between ground truth and detected results may still exist. But caused by the interaction across different keypoints, pose estimation typically fails more severely.

    \subsection{Cross Dataset Results}

\begin{table}[htbp]
\centering
\footnotesize
\begin{threeparttable}
\setlength{\tabcolsep}{2.8mm}{
\begin{tabular}{lcccc}
\toprule
\textbf{Test Set} & \textbf{HA(T)}  & \textbf{HA(F)} & \textbf{SK(T)}  & \textbf{SK(F)}\\
\midrule
\textbf{MSCOCO} &    63.5\tnote{\ddag}~ / 62.4\tnote{*}     &   74.5\tnote{\ddag}~ /{\color{HighlightRed} 74.8\tnote{*} }   &      13.5\tnote{*}   &  47.6\tnote{*} \\
\textbf{Human-Art} &  65.7 / { \color{HighlightRed}63.6\tnote{*}}   &   64.4 / 61.9\tnote{*}   &  14.3\tnote{*}  & 43.3\tnote{*} \\
\textbf{Sketch2Pose} &    \color{HighlightRed} 71.6\tnote{*}      &    70.5\tnote{*}     &     16.3\tnote{*}  & 68.9\tnote{*}  \\
\bottomrule 
\end{tabular}}
\begin{tablenotes}
            \scriptsize
            \item[1] \textbf{Training Strategies}: \textbf{HA} for \DatasetName~. \textbf{SK} for Sketch2Pose. \textbf{T} for training from scratch. \textbf{F} for fine-tuning from the HRNet pretrained on MSCOCO.
            \item[2] * / \ddag ~means calculating on the $10$ / $17$ intersected keypoints of different datasets (\DatasetName~\& Sketch2Pose / \DatasetName~\& MSCOCO).
    \end{tablenotes}
    \end{threeparttable}
\caption{AP results of pose estimation model HRNet on $3$ test sets (the first column) under $4$ training strategies (the first row). The best results on $10$ intersected keypoints are shown in {\color{HighlightRed} Red}.}
\label{tab:rebuttal_train_finetune}
\vspace{-0.5cm}
\end{table}

Due to the page limit, we put cross-dataset experimental evaluation on \DatasetName~ and Sketch2Pose in Table~\ref{tab:rebuttal_train_finetune}.  MSCOCO, Sketch2Pose, and \DatasetName~ have different keypoint definitions, thus we give out results on the 10/17 intersected keypoints of the three datasets (the three datasets have 10 intersected keypoints. MSCOCO and \DatasetName~ have 17 intersected keypoints). Four training setting is shown in the table, where (1) HA(T) means using \DatasetName~ for training, (2) HA(F) means using \DatasetName~ to finetune the model pre-trained on MSCOCO, (3) SK(T) means using Sketch2Pose for training, (4) SK(F) means using Sketch2Pose to finetune the model pre-trained on MSCOCO. Results show that Sketch2Pose is not sufficient for the training of multi-scenario pose estimator and thus shows poor results. Both training and fine-tuning with \DatasetName~ can lead to a relatively satisfactory accuracy, where fine-tuning with \DatasetName~ has the highest AP in the natural scenario, and training with \DatasetName~ has the best results on \DatasetName~and Sketch2Pose. Considering the commonness of natural humans in daily life, the recommended usage of \DatasetName~ is still combining training \DatasetName~ with MSCOCO.

\section{Datasets for Multi-Scenario Generalization}
    \label{sec:related_work_multi_scenario_datasets}

    Existing datasets~\cite{inoue2018cross,venkateswara2017deep,li2017deeper,westlake2016detecting,wu2014learning,Wilber_2017_ICCV,gonthier2018weakly} that include multi-scenario are more often used in domain generalization. 
    They focus on object classification or object detection tasks. Related methods try adapting classifiers or detectors from natural to artificial images\cite{Yao_2022_CVPR,xu2022h2fa}. 
    However, as demonstrated in Table \ref{tab:multi-scenario_datasets}, several limitations of these datasets make them hard to bridge natural and artificial human-centric tasks.
    First, the number of images and categories in these datasets is insufficient.
    Second, the number of downstream tasks they can support is constrained by the fact that they only have bounding box or object category annotations. Third, these datasets contain only a small percentage of scenes with humans and are not applicable to human-centric tasks.
    Besides, BAM!~\cite{inoue2018cross} is a large-scale dataset with 7 artificial categories targeted at image classification, but it uses untrustworthy model classifiers to label images instead of manually labeling, which may result in a lot of labeling mistakes.




    
    \begin{table}[htbp]
    \centering
    \small
    \begin{threeparttable}
    \setlength\tabcolsep{3pt} 
    \begin{tabular}{ccccc}
    \toprule
    \rowcolor[HTML]{EFEFEF} 
    \textbf{Task}                                                                                 & \textbf{Dataset}                                              & \multicolumn{1}{l}{\cellcolor[HTML]{EFEFEF}\textbf{Image}} & \textbf{\begin{tabular}[c]{@{}c@{}}Natural\\ Scenario\end{tabular}} & \textbf{\begin{tabular}[c]{@{}c@{}}Artificial\\ Scenario\end{tabular}} \\ \midrule
    
    \multicolumn{1}{l|}{\cellcolor[HTML]{EFEFEF}}                                                 & Inoue N. et al.~\cite{inoue2018cross}   & 5,000                                                      & 1                                                                  & 3                                                                     \\
    \multicolumn{1}{l|}{\cellcolor[HTML]{EFEFEF}}                                                 & Office-Home~\cite{venkateswara2017deep} & 15,500                                                     & 2                                                                  & 2                                                                     \\
    \multicolumn{1}{l|}{\cellcolor[HTML]{EFEFEF}}                      & PACS~\cite{li2017deeper}                & 9,991                                                      & 1                                                                  & 3                                                                     \\
    \multicolumn{1}{l|}{\cellcolor[HTML]{EFEFEF}}     
    & People-Art\cite{westlake2016detecting}         & 1,490                                                       & 1                                                                  & 42 \tnote{*}                                                                    \\
    \multicolumn{1}{l|}{\cellcolor[HTML]{EFEFEF}}                                                 & Photo-Art\cite{wu2014learning}         & 5,375                                                       & 1                                                                  & 1                                                                     \\
        \cline{2-5}
    \multicolumn{1}{l|}{\multirow{-5}{*}{\cellcolor[HTML]{EFEFEF}\rotatebox{90}{\begin{tabular}[c]{@{}c@{}}\textbf{Object}\\\textbf{Classification} \\ / \textbf{Detection}\end{tabular}}}}
    &\textbf{\DatasetName~(Ours)}                                               & \textbf{50,000}                                                     & \textbf{4}                                                                  & \textbf{16}                                                                    \\ \bottomrule
    \end{tabular}
        \begin{tablenotes}
            \footnotesize
            \item[*] The 42 painting styles of People-Art\cite{westlake2016detecting} have different classification criteria from \DatasetName, and these styles are encapsulated within the proposed 20 categories of \DatasetName.
    \end{tablenotes}
    \end{threeparttable}
    \caption{Comparison of multi-scenario datasets that serve for general object classification and detection tasks.}
    \label{tab:multi-scenario_datasets}
    \end{table}

    By contrast, \DatasetName~ is a full-scenario human-centric dataset inclusive of domain generalization tasks of both previous multi-scenario datasets such as human detection domain generalization, and other tasks such as human pose estimation, image generation, and image style transfer.

    Previous methods solve domain gap problems of object detection by transferring knowledge from the source domain to the target domain. \cite{westlake2016detecting} fine-tune Faster R-CNN on People-Art to detect humans in artworks. H2FA R-CNN~\cite{xu2022h2fa} proposes a Holistic and Hierarchical Feature Alignment R-CNN to enforce image-level alignment for object detection. \cite{inoue2018cross} use image-level domain transfer and pseudo-labels from the source domain to train object detector SSD300\cite{liu2016ssd}. 
    
    Previous works~\cite{zhou2017towards,jiang2021regressive} have explored domain generalization and adaptation for human keypoint detection in the natural scenario. However, to the best of our knowledge, no previous works involve multi-scenario human keypoint detection in both natural and artificial scenes. 
    
    In a word, no suitable domain adaptation and domain generalization method in the literature can be directly applied to \DatasetName~ and we leave it to future work.

\clearpage
    
{\small
\bibliographystyle{ieee_fullname}
\bibliography{PaperForReview}
}

  \end{document}